\documentclass[a4paper,fleqn]{cas-dc}

\usepackage[numbers]{natbib}
\usepackage{caption}
\usepackage{amsmath,graphicx}
\usepackage{amssymb}
\usepackage{subcaption}
\usepackage{amsfonts}
\usepackage{comment}
\usepackage{algorithm,algpseudocode}

\usepackage{todonotes}
\graphicspath{{figures/}}
\usepackage[export]{adjustbox}

\begin{document}
\let\WriteBookmarks\relax
\def\floatpagepagefraction{1}
\def\textpagefraction{.001}
\shorttitle{Deep Probabilistic models}
\shortauthors{Weizhu Qian et~al.}

\title [mode = title]{Supervised and Semi-supervised Deep Probabilistic Models for Indoor Positioning Problems}                      

\author[1]{Weizhu Qian}[orcid=0000-0002-2291-4028]
\cormark[1]
\ead{E-mail address: weizhu.qian@utbm.fr}
\address[1]{CIAD, Université Bourgogne Franche-Comt\'e UTBM, 90010, Belfort, France}

\address[2]{Mosel LORIA UMR CNRS 7503, Université de Lorraine, 54506, Vandœuvre-lès-Nancy, France} 

\author[1]{Fabrice Lauri}
\author[1,2]{Franck Gechter}

\begin{abstract}
WiFi fingerprint-based indoor localization has been a popular research topic recently. In this work, we propose two novel deep learning-based models, the convolutional mixture density recurrent neural network and the variational autoencoder-based semi-supervised learning model. The convolutional mixture density recurrent neural network is designed for indoor next location prediction, in which the advantages of convolutional neural networks, recurrent neural networks and mixture density networks are combined. Furthermore, since most of real-world WiFi fingerprint data are not labeled, we devise the variational autoencoder-based model to compute accurate user location in a semi-supervised learning manner. Finally, in order to evaluate the proposed models, we conduct the validation experiments on two real-world datasets. The final results are compared to other existing methods and verify the effectiveness of our approaches.  

\end{abstract}



\maketitle

\section{Introduction}
\label{sec:Intro}

Location-based services (LBS) are essential for applications such as location-based advertising, outdoor/indoor navigation, social networking, etc. Studying human mobility is important for developing LBS applications. Meanwhile, with the help of significant advances of the smartphone technology in recent decades, smartphone devices are integrated with various built-in sensors, such as Global Positioning System (GPS) modules, WiFi modules, cellular modules, etc. Acquiring the data from such kinds of sensors enables researchers to study human activities. Therefore, as opposed to some other approaches which are focusing on using base stations or radars, like RADAR~\cite{bahl2000radar} and HORUS~\cite{youssef2008horus}, we choose to base our work on the data collected from smartphones.

There are several types of smartphone data that can be used for such research. For instance, GPS equipment can provide the position information with latitudes and longitudes directly when smartphone users stay outdoors. The GPS-based localization methods are favored by many researchers~\cite{cho2016exploiting},~\cite{yu2017modeling}. However, this type of methods are not suitable for indoor positioning tasks because GPS modules do not work well in indoor environment.

In some cases, we have to use other indirect data to interpret the location information. Since WiFi connections are now widely used, one widespread approach is based on the detected WiFi fingerprints of smartphone devices. In this case, the received signal strength indicator (RSSI) values of WiFi access points (WAPs) scanned by mobile phones are used to compute the locations of users instead of using latitudes and longitudes directly. As compared to GPS-based localization methods, WiFi fingerprints-based localization methods can work in indoor environment as well.

In our work, we attempt to interpret WiFi fingerprints (RSSI values) into accurate users location (coordinates). However, this task is not easy to accomplish. Typically, the RSSI values data is composed of vectors whose elements represent the unique WAP IDs. To provide the good quality of WiFi connection, modern public buildings usually are equipped with a relatively large number of WiFi access points. As a result, this leads to the high dimensionality problem. Another issue is that RSSI values are not always stable due to the signal-fading effect and multi-path effect~\cite{hoang2019recurrent}. For example, a virtual WAP presents itself as multiple access points but in fact it only has one physical access point. In addition, according to our investigation, we find that an ordinary deep learning regression model with Euclidean distance-based loss is not powerful enough to overcome such difficulties.

In this paper, we aim at addressing two WiFi fingerprint-based positioning issues. This first problem is to predict the next time point user location with current WiFi fingerprints. The second problem is to recognize user location by semi-supervised learning. Accordingly, we propose two deep learning-based models to tackle these two problems respectively. The first proposed model is called the convolutional mixture density recurrent neural network (CMDRNN) designed to predict user paths. In the CMDRNN model, to tackle the high dimensionality issue, we deploy a one-dimensional convolutional neural network as a sub-structure to detect the features of the input. Moreover, in order to overcome the instability problem of the data, a mixture density network sub-structure is incorporated into our model for sampling the final output. Finally, since our task is a time-series prediction task, we model the state transition via a recurrent neural network sub-structure. With such unique design, the CMDRNN model is able to predict user location with WiFi fingerprints.

As we know, labelling data usually is both time-consuming and labor-consuming, thus most of real-word data in fact is unlabeled. However, even so, we still want to make as much use of accessible data as possible. To this end, as the second proposed approach, we propose a deep learning-based model, the variational autoencoder-based semi-supervised learning model. In this approach, we assume that the input (RSSI values) and the target (user location) are related to the same latent distribution. Therefore, in the unsupervised learning process, we use the variational autoencoder model to learn the latent distribution of the input whose information is relatively abundant. Then, in the supervised learning process, the labeled data is used to train the predictor. In this way, we can exploit more information of the dataset than other supervised learning methods.

The main contributions of our work are summarized as follows:

\begin{itemize}
    \item We devise a novel hybrid deep-learning model (CMDRNN) which allows us to predict the accurate position of smartphone users based on WiFi fingerprints;  
    
    \item We devise a variational autoencoder-based semi-\\supervised learning model for accurate indoor positioning using WiFi fingerprints;
    
    \item We conduct the evaluation experiments on two real-world datasets and compare our methods with other existing machine learning and deep learning methods.  

\end{itemize}

The reminder of the paper is organized as follows. Section~\ref{sec:Related} surveys the related work. Section~\ref{sec:Problem} describes the problems we will solve in this paper. In Section~\ref{sec:Method}, the two proposed methods are introduced. Section~\ref{sec:Experiments} presents the validation experiments and the results using the real-world datasets. Finally, we draw some conclusions and discuss about the potential future work in Section~\ref{sec:Conclusions}.

\section{Related Work}
\label{sec:Related}

In literature, researchers have explored various types of machine learning techniques, both conventional machine learning and deep learning methods, on location recognition and prediction with WiFi fingerprints data. 

\subsection{Conventional Machine Learning Methods}

In the work of~\cite{bozkurt2015comparative}, authors compared many traditional machine learning methods, decision trees (DTs), k-nearest neighbors (k-NNs), naive Bayes (NBs), neural networks (NNs), etc., for classifying buildings, floors and regions. In~\cite{cramariuc2016clustering}, authors clustered the 3D coordinates data by K-means and clustered the RSSI data by the affinity clustering algorithm, respectively. In particular, in~\cite{torres2015comprehensive}, researchers compared $51$ different distance metrics to investigate the most suitable distance functions for accurate WiFi-based indoor localization. Some researchers used Gaussian processes (GPs)~\cite{rasmussen2003gaussian} to model the relationship between WiFi signal strength and indoor location~\cite{ferris2007wifi},~\cite{hahnel2006gaussian},~\cite{yiu2015gaussian}. But GPs are not scalable to large datasets because of the expensive computational cost. In addition, for representing the time series state transition, due to the complex relationship between WiFi RSSI values and user coordinates, neither Kalman filter-based approa\-ches \cite{yang2019online} nor hidden Markov model-based approaches~\cite{krogh2001predicting} is suitable for our tasks .

\subsection{Deep Learning Methods}

Deep learning methods, such as convolutional neural networks (CNNs)~\cite{lecun1998gradient}, autoencoders (AEs)~\cite{hinton2006reducing} and recurrent neural networks (RNNs)~\cite{elman1990finding} also have been utilized in WiFi-based positioning tasks. For instance, \cite{ibrahim2018cnn} and \cite{ayyalasomayajula2020deep} proposed the CNN-based models for indoor localization. Generally, a buildings has many different WiFi access points, consequently the RSSI data in many situations, can be very high dimensional. For this reason, it is reasonable to reduce the input dimension before carrying out a regression task or a classification task. Some deep learning-based dimension-reduction methods like autoencoders can be the appropriate approaches~\cite{nowicki2017low},~\cite{song2019novel},~\cite{kim2018scalable}. For example, in the research of~\cite{song2019novel}, authors used an autoencoder network to reduce the data dimension, then used a CNN to proceed accurate user positioning. In~\cite{nowicki2017low},~\cite{kim2018scalable}, authors used autoencoders to reduce the input dimension before using a multi-layer perceptron (MLP) to classify the buildings and floors.  

In terms of time series prediction, in~\cite{hoang2019recurrent}, authors compared different types of recurrent neural networks including the vanilla RNN, the long short-term memory (LSTM), the gated recurrent unit (GRU) and the bidirectional LSTM for accurate RSSI indoor localization. They also employed a weighted filter for both input and output layers to enhance the sequential modeling accuracy. 

\subsection{Limitations of Conventional Neural Networks}

As for traditional neural networks (NNs), they can be regarded as deterministic models once trained, which can be described as follow:
\begin{equation}\label{Eq:NN_Eq1}
y = \mathcal{F}(x;w)
\end{equation}

where $x$ and $y$ are the input and output of the NN, respectively; $F$ represents the neural network structure and $w$ are the weights of the NN.   

Accordingly, the training loss (for instance, mean squared errors) of NNs can be described as follow:
\begin{equation}\label{Eq:NN_Eq2}
Loss = \frac{1}{N}\sum_{n=1}^{N}{(\hat{y_n}-y_n)^2}
\end{equation}

where $N$ is the mini batch size and $\hat{y}$ is the target.

In many situations, a NN model is powerful enough to obtain satisfying results. However, in some cases, for instance, a high non-Gaussian inverse problem, traditional neural networks will lead to very poor modeling results~\cite{bishop2006pattern}. A good solution to this issue is to seek a framework that can calculate conditional probability distributions between the input and the target.

Mixture density networks (MDNs) solve this problem by using maximum likelihood estimation (MLE)~\cite{bishop1994mixture}. In MDNs, the final output are sampled by mixture distributions rather than computed directly. One advantage of MDNs is that it can be applied to an estimation situation in which a large variety lies. For instance, we can incorporate more mixture of Gaussians to a MDN to enhance its estimating capacity for more complex distributions. However, as a MLE approach, the MDN also has obvious disadvantages. First, it needs to set some hyper-parameters properly (i.e., the number of Gaussians for a MDN), otherwise, it may not be able to provide desirable results. Moreover, MLE can be biased when samples are small, thus MDNs are not suitable for semi-supervised learning. In practice, we find that MDNs also suffer from computational instability when the mixture number is too large.

To alleviate the disadvantages of MDNs, we can resort to maximum a posterior (MAP)-based models, in which the prior of model parameters and the likelihood are both considered. MAP has the regularizing effect which can be used to prevent overfitting. For instance, Bayesian neural networks (BNNs), which apply Bayesian inference, have been introduced~\cite{hernandez2015probabilistic}. Unfortunately, in practice, we find that BNNs are not flexible enough for very complex distributions like our cases. 

To solve this problem, we use another Bayesian deep learning model, variational autoencoders (VAEs)~\cite{kingma2014auto}, in the proposed model so as to introduce a richer prior information. Meanwhile, since variational autoencoders are unsupervised deep latent generative models, we can use it to learn latent representation from the unlabeled data and to devise the semi-supervised learning model.

\section{Problem Description}
\label{sec:Problem}

In this work, our purpose is to calculate smartphone user location through using corresponding WiFi fingerprints. As opposed to some previous work, we attempt to obtain accurate user location, namely, coordinates (either in meters or longitudes and latitudes) rather than treat this subject as a classification task (identifying buildings or floors).The main goals of the two tasks in our work are summarised as follows: 


\begin{itemize}
    \item In the first task, we aim to predict the next indoor location of the users. The input is the WiFi RSSI values at the current time points and the target is the coordinates of the users at the next time points. This task can be regarded as a high dimensional time-series regression task. In the later section, we will introduce a hybrid deep learning model to solve this problem; 
    
    
    \item In the second task, we aim to compute the location of the users. The input is the WiFi RSSI values and the target is the coordinates of the users. We do not treat the data as sequences. This task can be regarded as a high dimensional regression task. In particular, we wan to investigate semi-supervised learning methods to solve this problem, therefore we split the input data into the labeled set and unlabeled set. In the later section, we will introduce a semi-supervised deep learning model to solve this problem.
    
\end{itemize}

\begin{table*}
\small
\centering
\caption{Descriptions for the two tasks}
\label{Tab:Tasks}
\begin{tabular}{|c|c|c|c|c|c|c|c|}
\hline
Task & Input & Output & Learning method & Purpose\\
\hline
Task 1 & RSSI values (sequential) & coordinates (sequential) & supervised & location prediction\\
Task 2 & RSSI values & coordinates & semi-supervised & location recognition\\

\hline
\end{tabular}
\end{table*}

Table~\ref{Tab:Tasks} describes the differences between the two tasks. To accomplish these two tasks, we introduce two novel deep learning-based probabilistic models, the convolutional mixture density recurrent neural network and the variational \\ autoencoder-based semi-supervised learning model in this paper.

\section{Proposed Methods}
\label{sec:Method}

As explained before, we introduce two proposed deep learning-based models in this section. The first proposed model is called the convolutional mixture density recurrent neural network (CMDRNN), which is designed to predict paths with WiFi fingerprints. The second proposed model is the VAE-based model, which is designed to proceed semi-supervised learning for WiFi-based positioning.    

\subsection{Convolutional Mixture Density Recurrent Neural Network}

\subsubsection{1D Convolutional Neural Network}

In our first task, the input features are composed of the RSSI values of all the WiFi access points (WAPs) in the buildings, therefore the input dimension can be very high. Since the features of the WiFi fingerprint data represent the different WiFi WAP IDs. The adjacent features suggest that they are close spatially in the real world. Therefore, their RSSI values are more similar when the users are approaching compared to the WAPs that are remote to them (it will be illustrated in the WiFi data samples in the experiments). For this reason, to deal with the high dimensionality problem, we choose to use a convolutional neural network~\cite{lecun1998gradient}. CNNs are powerful tools for detecting features and widely used for tasks such as image processing, natural language processing and sensor signal processing. In particular, since the input of our model are RSSI value vectors, we adopt the 1D convolutional neural network to extract the properties of the high dimensional input.

\subsubsection{Recurrent Neural Network}

Recurrent neural networks (RNNs)~\cite{elman1990finding} are widely used for natural language processing (NLP), computer vision and other time series prediction task. In our model, we employ a RNN model to forecast the user location. 

The state transitions of RNNs can be expressed as follow:   
\begin{equation}\label{Eq:RNN_state}
h_t =\sigma_h(W_h{x_t}+{U_h}h_{t-1}+b_h)
\end{equation}
where $x_t$ is the input, $h_t$ is the hidden state, $\sigma_h$ is the activation function, $W_h$ are the weights from the input layer to the hidden layer, $U_h$ are the weights in the hidden layers and $b_h$ are the biases.

The output of a conventional RNN can be described as follow: 
\begin{equation}\label{Eq:RNN_output}
y_t =\sigma_y(W_y{h_t}+b_y)
\end{equation}

where $y_t$ is the output, $\sigma_y$ is the activation function, $W_y$ are the weights and $b_y$ are the output biases.

However, in many situations, RNNs may suffer from the long-term dependency problem during the learning process. To address this issue, the researchers proposed a variant of RNNs, called the long short-term memory network (LSTM)~\cite{hochreiter1997long}. In a LSTM, each unit has three gates, namely, an input gate, an output gate and a forget gate. These gates regulate the cell states of the LSTM to avoid the long-term dependency problem. More recently, the researchers proposed a variant of RNN, called the gated recurrent unit (GRU)~\cite{chung2014empirical}, which has the similar accuracy as LSTMs but less computing cost. We will deploy these three RNN architectures in the proposed model respectively for comparisons.     

\subsubsection{Mixture Density Network}

A traditional neural network with a mean squared error-based loss function, is optimized by a gradient-descent based method. Generally, such a model can perform well on the problems that can be described by a deterministic function $f(x)$, i.e., each input corresponds to an output of one specific value. However, for some stochastic problems, one input may has more than one possible values. Generally, this kind of problems are better to be described via a conditional distribution $p(y|x)$ than a deterministic function $y = f(x)$. In such cases, traditional neural networks may not work as expected.  

To address this issue, we can replace the original loss function with a conditional function, for a regression task, the Gaussian distribution can be a proper choice. Moreover, utilizing mixed Gaussian distributions can improve the representation capacity of the model. Based on this idea, researcher proposed mixture density networks (MDNs) model~\cite{bishop1994mixture}. In contrast with traditional neural network, the outputs of a MDN are the parameters of a set of mixed Gaussian distributions and the loss function becomes the conditional probabilities. Therefore, the optimization process is to minimize the negative log probability. Hence, the loss function can be described as follow:
\begin{equation}\label{Eq:MDN_output}
p(y|x)=\sum_{k=1}^{K}{\pi_k p(y |x; \theta_k)}
\end{equation}

where $x$ is the input, $K$ is the total mixture number, $\pi_k$ is the assignment probability for each model, with $\sum_{k=1}^{K}\pi_k=1, (0 \leq \pi_k \leq 1)$, and $\theta_k$ are the internal parameters of the base distribution; 
For Gaussians, $\theta_k =\{\mu_k,\sigma_k\}$, $\mu_k$ are the means and $\sigma_k$ are the variances.

Accordingly, in our proposed model, the original output layer of the RNN, Eq.~(\ref{Eq:RNN_output}), is rewritten as: 
\begin{equation}\label{Eq:CMDRNN_output}
\theta_t =\sigma_\theta(W_\theta{h_t}+b_\theta)
\end{equation}

where $\theta_t$ is the output of the RNN sub-model and also the input of the MDN sub-model, $\sigma_\theta$ is the activation function, $W_\theta$ are the weights and $b_\theta$ are the biases.

After the training process, we can use the neural network along with the mixed Gaussian distributions to represent the target distribution.  

\subsubsection{Proposed Model}

\begin{figure*}
\centering
\includegraphics[width= 0.8\linewidth]{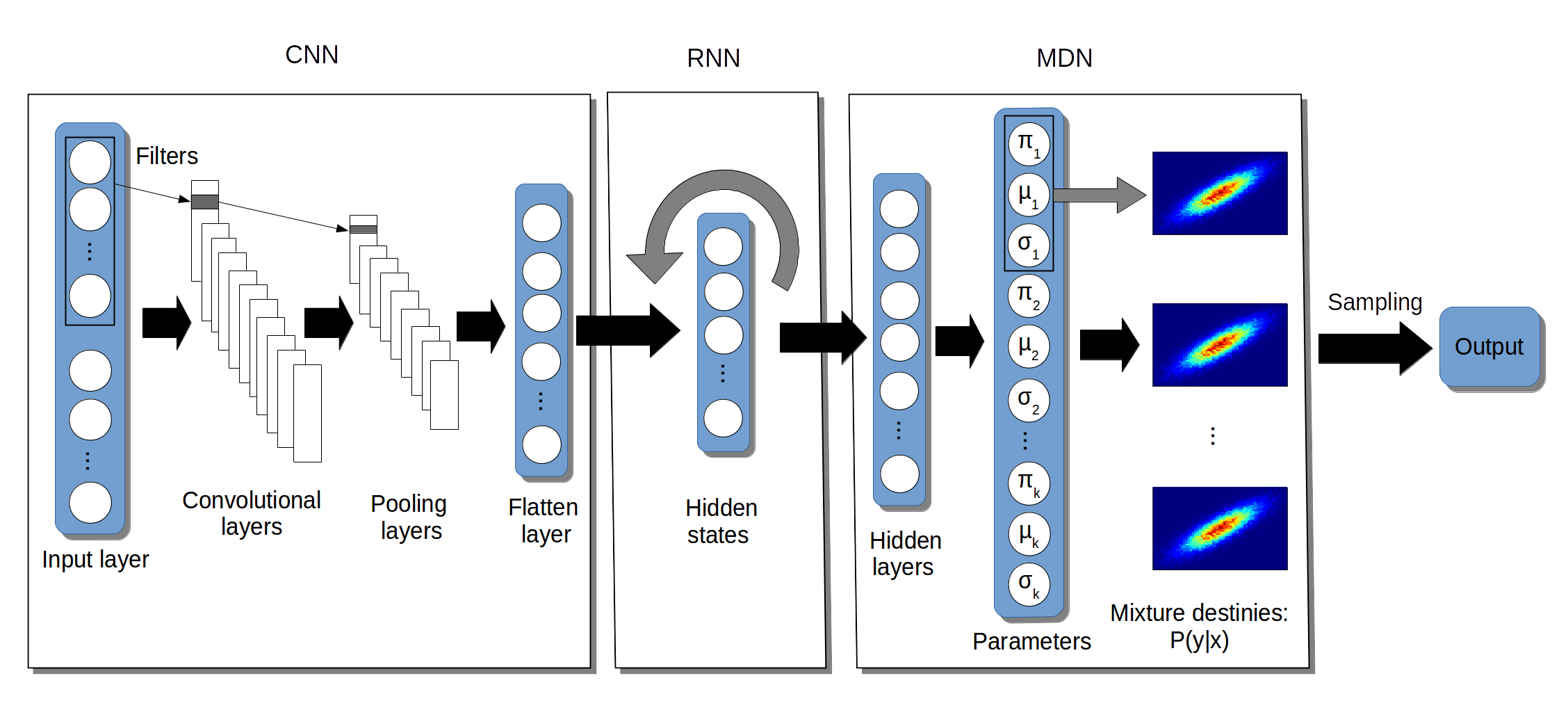}
\caption{Convolutional mixture density recurrent neural network.}\label{Fig:CMDRNN}
\end{figure*}

Knowing the merits of the three aforementioned neural networks, we devise a novel deep neural network architecture, called the convolutional mixture density recurrent neural network (CMDRNN). In the CMDRNN model, a 1D CNN is used to capture the features of the high dimensional input, then the state transitions of the time series data is modeled by a RNN model, and the output layer composed of mix Gaussian densities is used to sample the final prediction. With such a structure, we believe that our model is able to illustrate complex high dimensional time series data. Fig.~\ref{Fig:CMDRNN} shows the whole structure of the CMDRNN model and Algorithm~\ref{Alg:CMDRNN} demonstrates the learning process of the CMDRNN model. 

\begin{algorithm} 
\caption{Algorithm: the CMDRNN model}
\label{Alg:CMDRNN}
\begin{algorithmic}[1]
\Require{$x_t$ (RSSI Values)} 
\Ensure{$y_t$ (Coordinates)}
\Statex

\While{e $<$ max epoch}
    \While{ i $<$ batch num}
    
    \State{$h_0$ $\gets$ $Conv1d(x_t)$} \Comment{convolutional operation}
    \State{$h_1$ $\gets$ max pool $h_0$}
    \State{$f_t$ $\gets$ flatten $h_1$} 
    
    \State{$h_{t}$ $\gets$ $\sigma_h(W_h*f_t+U_h*y_{t-1}+b_h)$} \Comment{update hidden states}
    \State{$\theta_t$ $\gets$ $\sigma_y(W_y*h_t+b_y)$} \Comment{compute RNN model output}
    \State{$\theta_k$ $\gets$ $\theta_t$} \Comment{assign mixture density parameters }
    \State{minimize loss function: $-p(y_t|x_t;\theta_k)$}
   
    \EndWhile
\EndWhile
\State{$y_t \sim p(y_t|x_t;\theta_k)$}\Comment{sample final output} \\
\Return{$y_{t}$}

\end{algorithmic}
\end{algorithm}

The uniqueness of our method is that, compared with other existing models in literature, our model adopts a sequential density estimation approach. Thus, the learning target of the proposed method is a conditional distribution of the data rather than a common regressor. By doing so, our model can solve the complex sequential modeling task in this work. 

\subsection{VAE-based Semi-supervised Learning}

Now, we tackle the WiFi-fingerprint location recognition task via semi-supervised learning. To make the semi-supervised learning scheme applicable for our case, we need to make two important assumptions first: 
\begin{itemize}
    \item Assume that the input $x$ (RSSI values) and the target $y$ (coordinates) are related to the same latent variable $z$ (related to user position);
    
    \item Assume that a statistic $t(x)$ solely of $x$ is enough to be the sufficient statistic for $z$, which means $t(x)$ captures all the necessary information for calculating the parameters in the distribution of $z$. 
\end{itemize}

The first assumption describes the relationship among the input $x$, the output $y$ and the latent variable $z$. The second assumption explains why we can infer the distribution of the latent variable $z$ with the input $x$, which can be regarded as an unsupervised learning process. Since in many real-world cases, available datasets have more information about the input $x$ and less information about the target $y$, therefore it is more reliable to infer the latent distribution of $z$ via $p(z|x)$ rather than via $p(z|y)$.

Let $q(z)$ represent the prior distribution of $z$, then we can approximate $p(z|x)$ via variational inference. Thus, for the unsupervised learning process, our goal is to minimize the Kullback–Leibler (KL) divergence  $D_\text{KL}$.
\begin{equation}
\label{Eq:Met_Eq2}
D_{KL}\big( p(z|x) || q(z) \big) = \iint p(z|x) \log \frac{p(z|x)}{q(z)} dzdx
\end{equation}


According to the chain rule and the assumptions we made, then the generative procedure can be formulated as follow:
\begin{align}
\label{Eq:Met_Eq3}
p(y,z,x)  = p(y|z,x)p(z|x)p(x)
\end{align}

Accordingly, the predicting procedure (either deterministic or stochastic) can be described as follow: 
\begin{equation}
\label{Eq:Met_Eq4}
y \sim p(y|z,x)
\end{equation}

We can implement Eq.~(\ref{Eq:Met_Eq3}) and Eq.~(\ref{Eq:Met_Eq4}) via an unsupervised learning process and a supervised learning process, respectively. Hence, our method consists of two steps:   
\begin{itemize}
\item The first step (unsupervised learning): we employ a deep generative model to obtain the latent distribution $p(z|x)$;

\item The second step (supervised learning): we employ a MLP structure to predict the target $y$ using the learnt latent distribution.  
\end{itemize}

\subsubsection{Unsupervised Learning Procedure}

For the unsupervised learning, we adopt a variational autoencoder as the generative model to learn the latent distribution. Variational autoencoders (VAEs)~\cite{kingma2014auto} are deep latent generative models which adopt variational inference. Different to conventional autoencoders or other generative models, the latent representations of VAEs are continuous probabilistic distributions, which can be used to represent the real user coordinates.

In VAEs, the prior of the latent variable $z$, $q(z)$ can be regarded as a standard Gaussian distribution: 
\begin{equation}\label{Eq:VAE_Eq1}
z \sim \mathcal{N}(0,\mathbb{I})
\end{equation}

In order to obtain the posterior $p(z|x;\phi)$ via the VAE encoder neural network, $z$ is reparameterized using the following equation: 
\begin{equation}\label{Eq:VAE_repara}
z = \mu_z+ \sigma_z\odot\epsilon,  \;  \epsilon \sim \mathcal{N}(0;\mathbb{I})
\end{equation}

where $\mu_z$ is the mean of $z$, $\sigma_z$ is the variance of $z$, $\odot$ denotes the Hadamard product. 

The VAE decoder is used to reconstruct the original input. It can be described as follow: 
\begin{equation}\label{Eq:VAE_Eq2}
x' \sim p(x|z;\theta)
\end{equation}
where $x'$ is the reconstructed input and $\theta$ is the decoder parameters.

The loss function of the VAE can be written as follow:
\begin{align}
\label{Eq:VAE_loss}
\mathcal{L}(\theta, \phi, D) = \mathop{{}\mathbb{E}}{_{z \sim p(z|x)}} \big [-\log p(x|z;\theta) \big] \nonumber \\
+ D_{KL} \big(p(z|x;\phi)||q(z)\big)
\end{align}

Once $\mathcal{L}(\theta, \phi, D)$ is minimized, we can have the approximate posterior $p(z|x;\phi)$ for sampling the latent variable $z$.

\subsubsection{Deterministic Predictor (M1 Model)}
After the unsupervised learning procedure, we can acquire the latent distribution $z$. The next step is to carry out the supervised learning procedure to obtain the target $y$. To this end, we devise two predictor models, one is deterministic, the other is stochastic.

Now we build a deterministic predictor which consists of two predicting steps:

Step $1$: to obtain the means of latent variables.
\begin{equation}\label{Eq:M1_Eq1}
\mu_z =\mathcal{F}_\mu(x;\phi)
\end{equation}

where $\mathcal{F}_\mu(x;\phi)$ can be regarded as the encoder of the VAE.  

Step $2$: to obtain the final prediction based on the output of Step $1$. 
\begin{equation}\label{Eq:M1_Eq2}
y = \mathcal{F}_y(\mu_z;w) 
\end{equation}

where $\mathcal{F}_y(\mu;w)$ is a deterministic multi-layer perceptron model.
Consequently, the loss function is:
\begin{equation}\label{Eq:M1_loss}
\mathcal{L}(D;w) = \frac{1}{N}\sum_{n=1}^{N}{(\hat{y}_n-y_n)^2}
\end{equation}

where $\hat{y}_n$ is the labeled target. 


Compared to using the original data as the input directly, using the learnt latent distribution as the input of the predictor can reduce the noisy information of the original input.

\begin{algorithm} 
\caption{Algorithm: M1 model}
\label{Alg:M1}
\begin{algorithmic}[1]
\Require{$x_a$ (all input), $x_l$ (labeled input), $\hat{y}$ (labels)} 
\Ensure{$y$ (predictions)}
\Statex

\While{unsupervised learning}
    
    \State{$\mu_z$, $\sigma_z$ $\gets$ $E_\phi(x_a)$}  \Comment{$E_\phi(x_a)$: encoder network}
    \State{$z \sim \mathcal{N}(\mu_z, \sigma_z)$}  \Comment{sample latent codes}
    
    \State{$x_a'$ $\gets$ $p(x|z;\theta)$} \Comment{$x_a'$: reconstructed input}
    \State{minimize loss function $\mathcal{L}(\theta, \phi, D)$}\Comment{Eq.~(\ref{Eq:VAE_loss})}
   
\EndWhile\\

\While{supervised learning}
    \State{$\mu_z$ $\gets$ $\mathcal{F}_\mu(x_l;\phi)$}  \Comment{get latent codes}
    \State{$y$ $\gets$ $\mathcal{F}_y(\mu_z;w)$}    \Comment{get predictions}
    \State{minimize loss function $\mathcal{L}(D;w)$}   \Comment{Eq.~(\ref{Eq:M1_loss})}
    
\EndWhile\\
\Return{$y$}

\end{algorithmic}
\end{algorithm}

The training process of the M1 model is summarized in Algorithm~\ref{Alg:M1}. 

\subsubsection{Stochastic Predictor (M2 Model)}

Alternatively, in contrast with the M1 model, we can also devise a stochastic predictor, the M2 model. To this end, based on Eq.~(\ref{Eq:Met_Eq3}), we decompose the joint distribution as follow:
\begin{equation}\label{Eq:M2_Eq1}
p(y,z,x;w,\phi) = p(y|z,x;w)p(z|x;\phi)p(x) 
\end{equation}

where $p(z|x;\phi)$ is the encoder network parameterized by $\phi$ and $p(y|z;w)$ the predictor network parameterized by $w$; $p(x)$ can be approximated via empirically drawing the samples from the dataset.   

Based on Eq.~(\ref{Eq:M2_Eq1}), we can formulate a stochastic prediction model. However, since Eq.~(\ref{Eq:M2_Eq1}) cannot be computed explicitly, we can use Monte Carlo method to solve it by drawing the samples of $z$ and $y$. To this end, first, we draw the latent variables $z$ from the VAE encoder: 
\begin{equation}\label{Eq:M2_Eq3}
z \sim p(z|x;\phi)
\end{equation}

Then, we draw the predicted values $y$: 
\begin{equation}\label{Eq:M2_Eq4}
y \sim p(y|z,x;w)
\end{equation}

Hence, based on Eq.~(\ref{Eq:VAE_loss}) and Eq.~(\ref{Eq:M2_Eq1}), the loss function of the M2 model can be written as:
\begin{equation}\label{Eq:M2_elbo}
\begin{array}{l}
\mathcal{L}(D;\theta,\phi,w) = \mathop{{}\mathbb{E}}{_{z \sim p(z|x)}} \big [-\log p(y|z,x;w) \big] \\ 
 \; \; \; \; + \mathop{{}\mathbb{E}}{_{z \sim p(z|x)}} \big[-\log p(x|z;\theta)\big] + D_{KL} \big (p(z|x;\phi)||q(z) \big)
\end{array}
\end{equation}

In Eq.~(\ref{Eq:M2_elbo}), the first term represents the predictor, the second term represents the decoder and the last term represents the encoder. The second term and the last term can be optimized by the unsupervised procedure jointly in the first step. Afterwards, since $\phi$ and $\theta$ are trained, according to Eq.~(\ref{Eq:M2_elbo}), here we only need to optimize $p(y|z,x;w)$ in the second step. Thus, the loss function for training the predictor becomes:
\begin{equation}\label{Eq:M2_Loss_Eq1}
\mathcal{L}(D;w) = \mathop{{}\mathbb{E}}{_{z \sim p(z|x)}} \big [-\log p(y|z,x;w) \big ]
\end{equation}

We can use Monte Carlo sampling to solve this loss:  
\begin{equation}\label{Eq:M2_Loss_Eq2}
\mathcal{L}(D;w) \approx -\frac{1}{N}\sum_{n=1}^{N} \log p(y|z,x;w)
\end{equation}
where $N$ is the mini batch size. Assume that the likelihood function $p(y|z;w)$ is a Gaussian distribution with noise $\sigma_y$ which can be seen as a hyper-parameter. For the prediction, we draws multiple samples via the predictor and use their mean values as the final output.   

\begin{algorithm} 
\caption{Algorithm: M2 Model}
\label{Alg:M2}
\begin{algorithmic}[1]
\Require{$x_a$ (all input),$x_l$ (labeled input),$y_l$ (labels)} 
\Ensure{$y$ (predictions)}
\Statex

\While{unsupervised learning}
    
    \State{$\mu_z$, $\sigma_z$ $\gets$ $E_\phi(x_a)$}  \Comment{$E_\phi(x_a)$: encoder network}
    \State{$z \sim \mathcal{N}(\mu_z, \sigma_z)$}  \Comment{sample latent codes}
    
    \State{$x_a'$ $\gets$ $p(x|z;\theta)$} \Comment{$x_a'$: reconstructed input}
    \State{minimize loss function $\mathcal{L}(\theta, \phi, D)$}\Comment{Eq.~(\ref{Eq:VAE_loss})}
   
\EndWhile\\

\While{supervised learning}
    \State{$z \sim \mathcal{N}(\mu_z, \sigma_z)$}  \Comment{sample latent codes}
    \State{$y \sim p_y(z; w)$}    \Comment{sample predictions}
    \State{minimize loss function $\mathcal{L}(D;w)$} \Comment{Eq.~(\ref{Eq:M2_Loss_Eq2})}
    
\EndWhile\\

\Return{$y$}

\end{algorithmic}
\end{algorithm}

\begin{figure}
\centering
\includegraphics[width=1.2\linewidth,center]{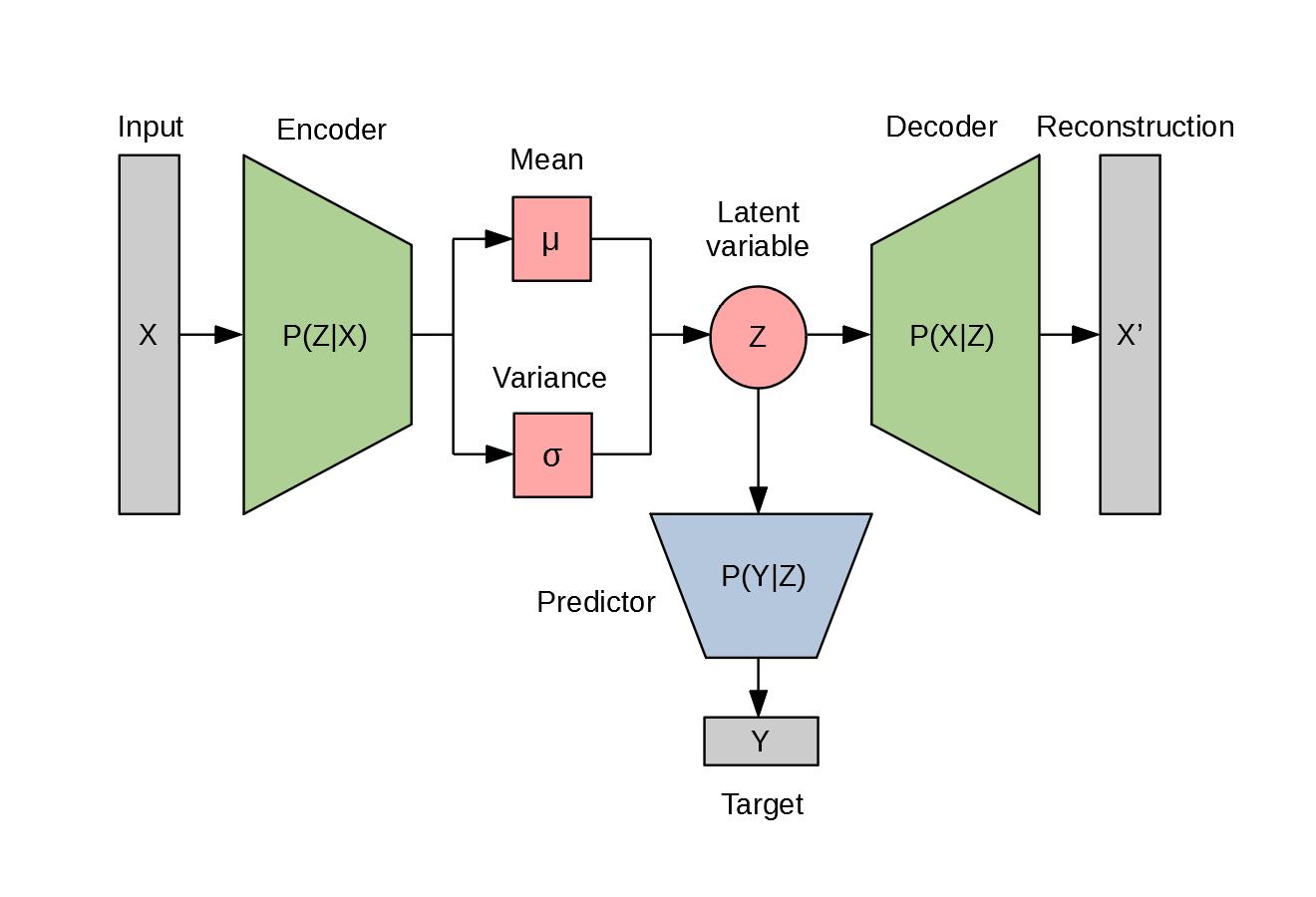}
\caption{VAE-based semi-supervised learning model.}\label{Fig:VAE_based_model}
\end{figure}

The training process of the M2 model is summarized in Algorithm~\ref{Alg:M2} and the structure of the VAE-based semi-supervised learning model is illustrated in Fig.~\ref{Fig:VAE_based_model}.

\section{Experiments and Results}
\label{sec:Experiments}
For the validation, we conduct two groups of experiments to evaluate the two proposed models, respectively. 

\subsection{Indoor Next Location Prediction}
\subsubsection{Dataset Description}

\begin{figure}
\centering
\includegraphics[width= 1.\linewidth]{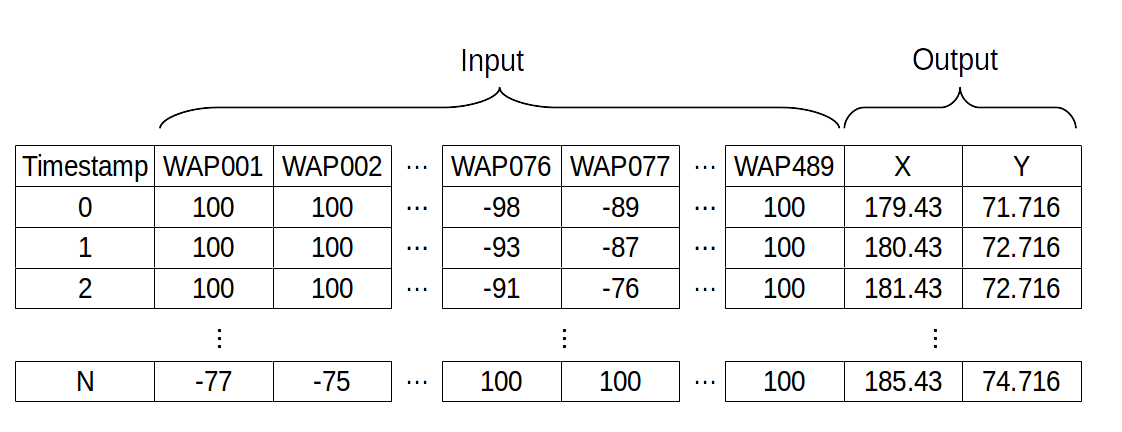}
\caption{WiFi fingerprint data samples.}
\label{Fig:Data}
\end{figure}

For the validation dataset, we select two paths from the Tampere dataset~\cite{lohan2017wi}. As shown in Fig.~\ref{Fig:Data}, the input dimension of the Tampere dataset is $489$. The possible RSSI values of the detected WAPs range from $-110 $ dB to $0$ dB and the RSSI values of undetected WAPs are set to $100$. The units of the target values are meters. For pre-processing the data, we set the undetected RSSI values to $0$ for the
purpose of computational convenience, which does not represent the actual signal strength. We only use the coordinates of the users to shape the paths. 

\subsubsection{CMDRNN Implementation Details}

The implementation details of our model are illustrated in Table~\ref{Tab:CMDRNN_Imp}. The CNN sub-network consists of three layers, a convolutional layer, a max pooling layer and a flatten layer. The RNN sub-network includes a hidden layer with $200$ neurons. The MDN sub-network has a hidden layer and an output layer. The mixed Gaussians number of the MDN output layer is $30$, and each mixture has $5$ parameters, namely, two dimensional means, diagonal variances and corresponding weights. We use RMSProp~\cite{tieleman2012lecture} as the optimizer.   

\begin{table*}
\centering
\caption{CMDRNN implementation details.}
\label{Tab:CMDRNN_Imp}
\begin{tabular}{|c|c|c|c|}
\hline
Sub-network & Layer & Hyperparameter & Activation function\\
\hline

CNN & convolutional layer & filter number: 100; stride: 2 & Sigmoid\\
CNN & max pooling layer & neuron number: 100 & ReLU \\
CNN & flatten layer & neuron number: 100  & Sigmoid \\
RNN & hidden layer & memory length: 5; neuron number: 200 & Sigmoid \\
MDN & hidden layer & neuron number: 200 & Leaky ReLU \\
MDN & output layer & 5*mixed Gaussian number (5*30) & - \\ 
\hline

\multicolumn{4}{|c|}{Optimizer: RMSProp; learning rate: $10^{-3}$} \\

\hline
\end{tabular}
\end{table*}

\subsubsection{Experimental Results}

\begin{figure}
\centering
    \begin{subfigure}[t]{0.45\textwidth}
    \includegraphics[width= 1.\linewidth]{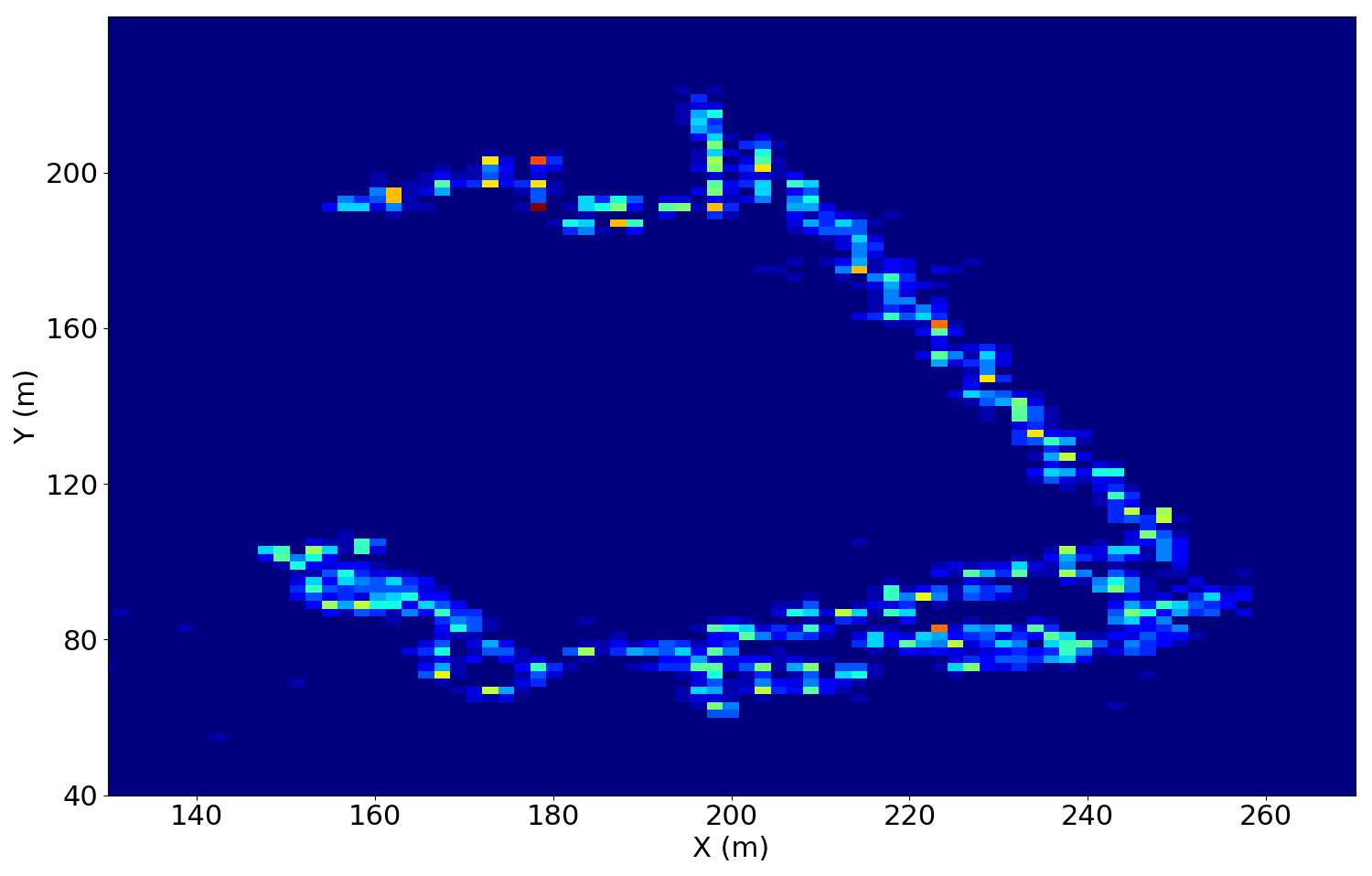}
    \caption{Path 1 prediction results.}
    \end{subfigure}

    \begin{subfigure}[t]{0.45\textwidth}
    \includegraphics[width= 1.\linewidth]{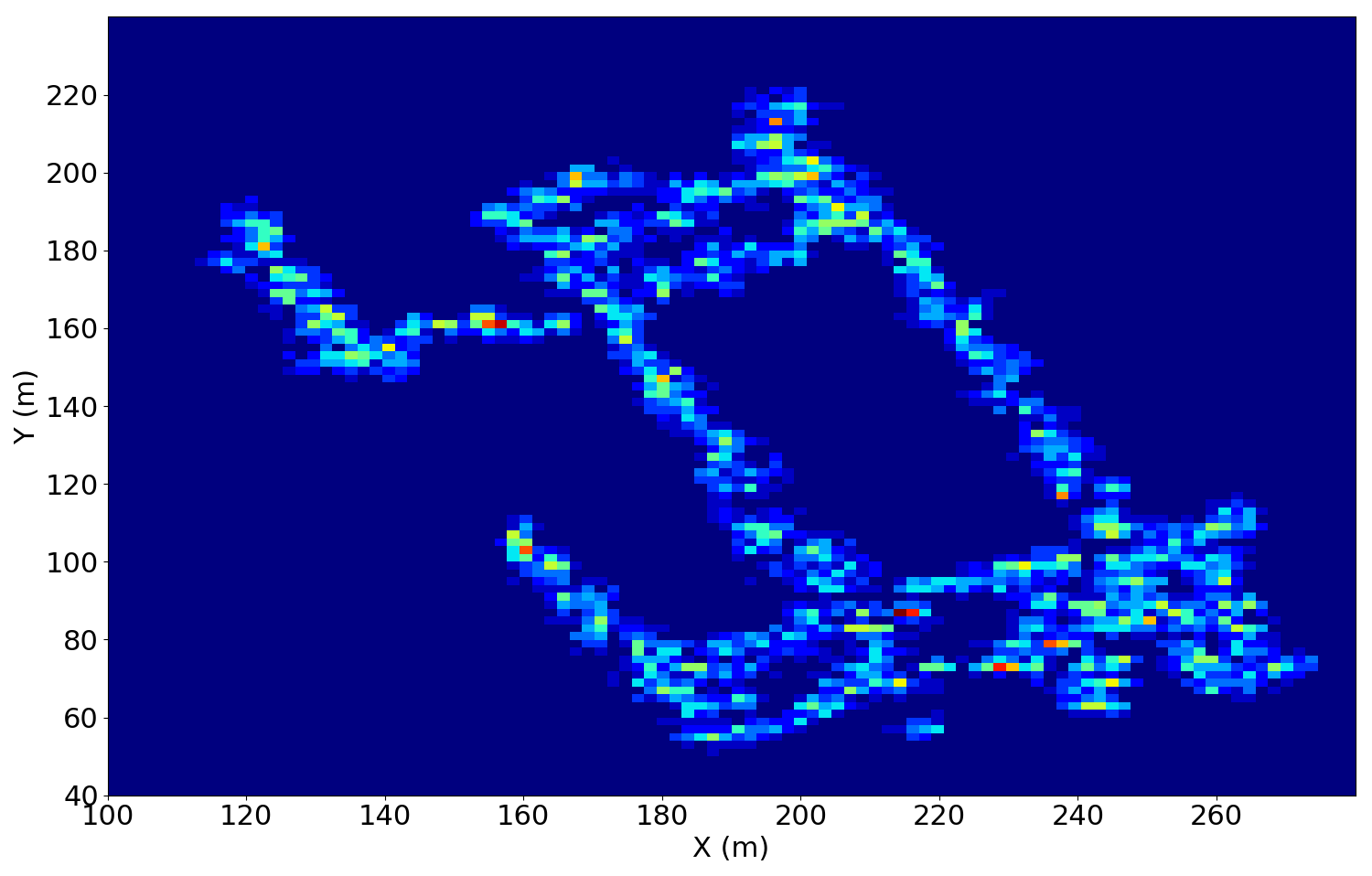}
    \caption{Path 2 prediction results.}
    \end{subfigure}
\caption{Prediction results of two selected paths.}
\label{Fig:Path_result}  
\end{figure}

In order to evaluate the effectiveness of our method, we conduct several experiments to thoroughly compare our CMDRNN model to other deep learning approaches. Fig. \ref{Fig:Path_result} demonstrates the prediction results of the two selected paths. 

\begin{figure}
\centering
\includegraphics[width= 1.\linewidth]{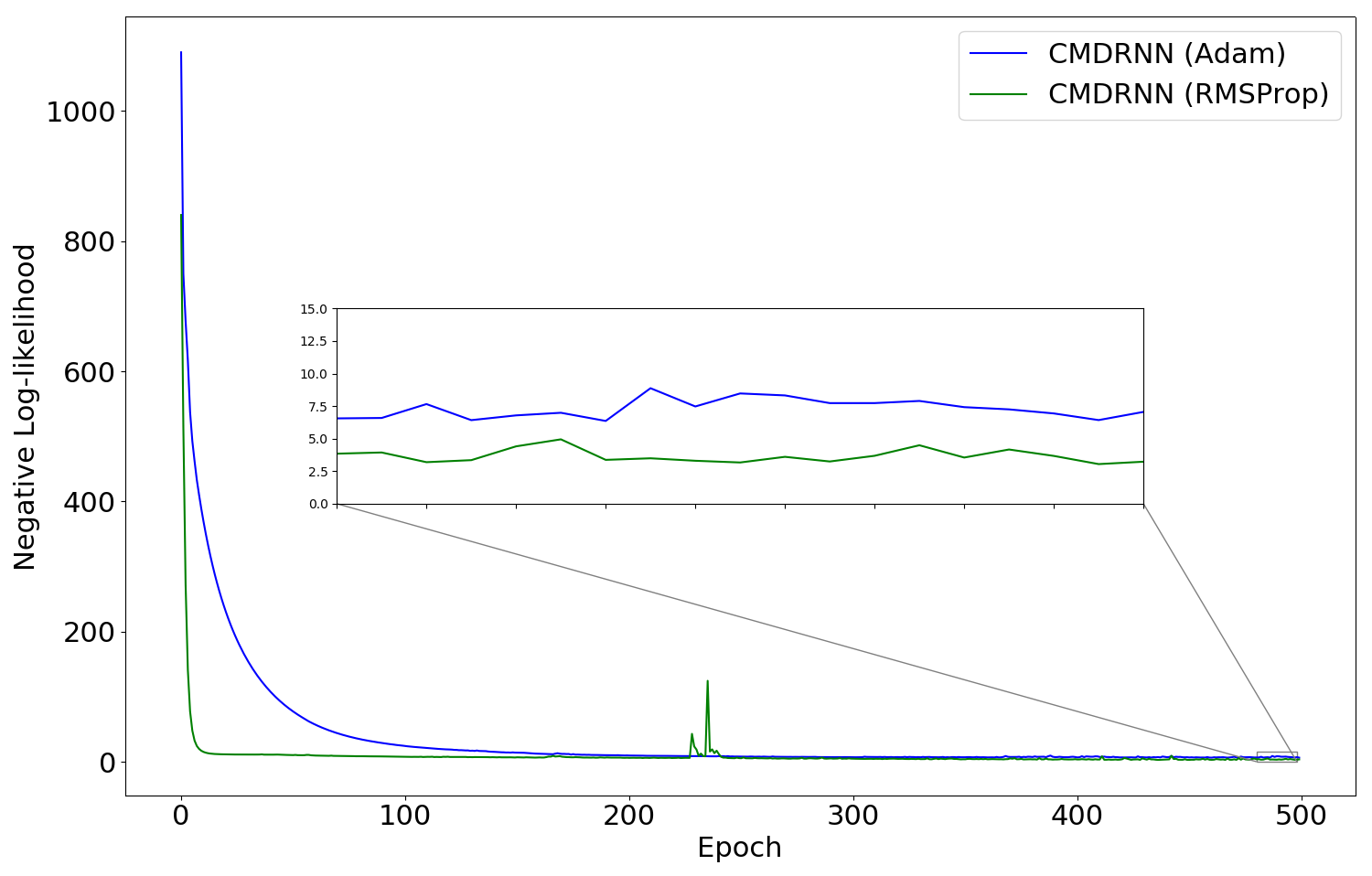}
\caption{Training losses using different optimizers.}\label{Fig:Loss_l}
\end{figure}

In the experiments, we test different optimizers and feature detectors and use learning losses as the metrics to evaluate the performances. In~\cite{martin2017wasserstein}, it reports that RMSProp~\cite{tieleman2012lecture} may have better performance on very non-stationary tasks than the Adam optimizer~\cite{kingma2015adam}. To verify this, we train our algorithm with RMSProp and Adam, respectively. As it is shown in Fig.~\ref{Fig:Loss_l}, the proposed model can converge to a lower negative log-likelihood using RMSProp than using Adam. Thus, we choose RMSProp as the optimizer to learn our model.    

\begin{figure}
\centering
\includegraphics[width= 1.\linewidth]{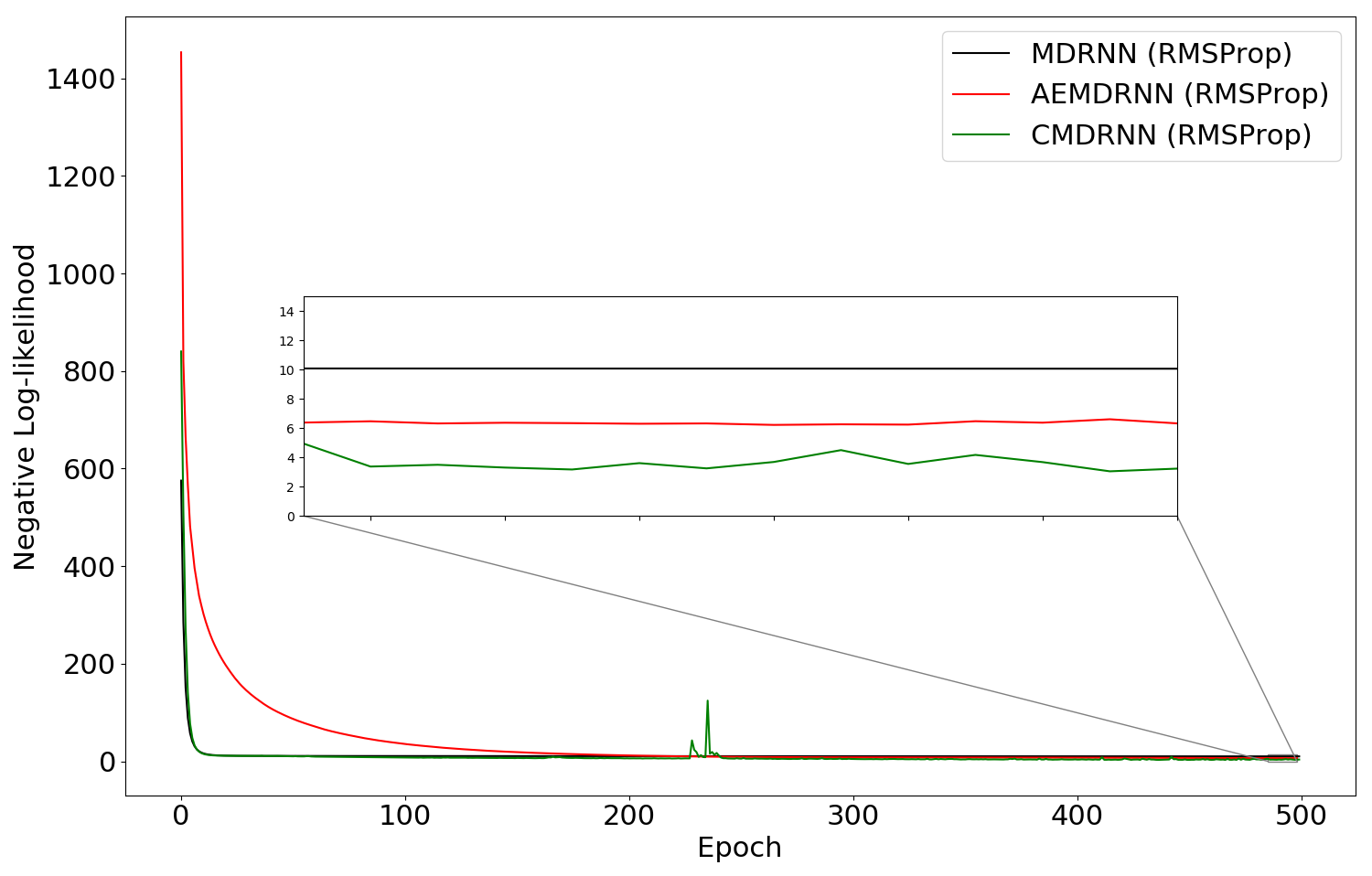}
\caption{Training losses using different feature detectors.}\label{Fig:Loss_f}
\end{figure}

In terms of feature detectors, since the input is high dimensional, a sagacious way to deal with this issue is incorporating a sub-network into the model for dimension reduction or feature detection. Many previous research adopted autoencoders to reduce dimension, while we argue that the more appropriate choice for the task in our work is using a one-dimensional CNN. In order to prove that, we test three different models, one without a feature-detecting structure, one using an autoencoder and one using 1D CNN (the proposed model). The autoencoder model with structure \{hidden neurons: $256$; hidden neurons: $128$; code size: $64$; hidden neurons: $128$; hidden neurons: $256$\}. As the results demonstrated in Fig.~\ref{Fig:Loss_f}, the proposed model suing the 1D CNN feature detector can reach lower negative log-likelihood during the training process. 

\begin{figure}
\centering
\includegraphics[width= \linewidth]{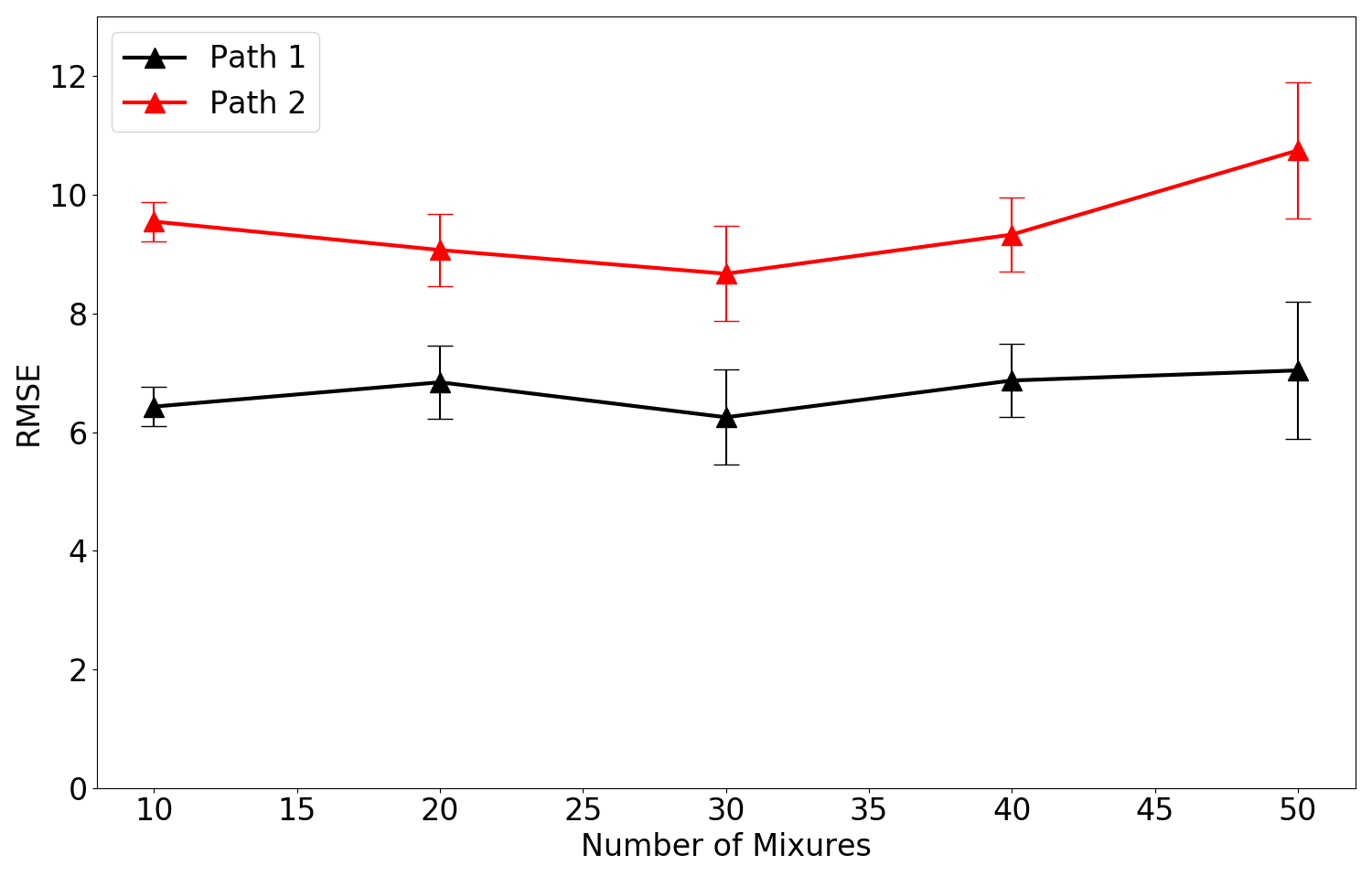}
\caption{Prediction results of different mixture numbers in the MDN (the bars represent the standard deviations).}\label{Fig:mixture_number}
\end{figure}

\begin{figure}
\centering
\includegraphics[width= 1.\linewidth]{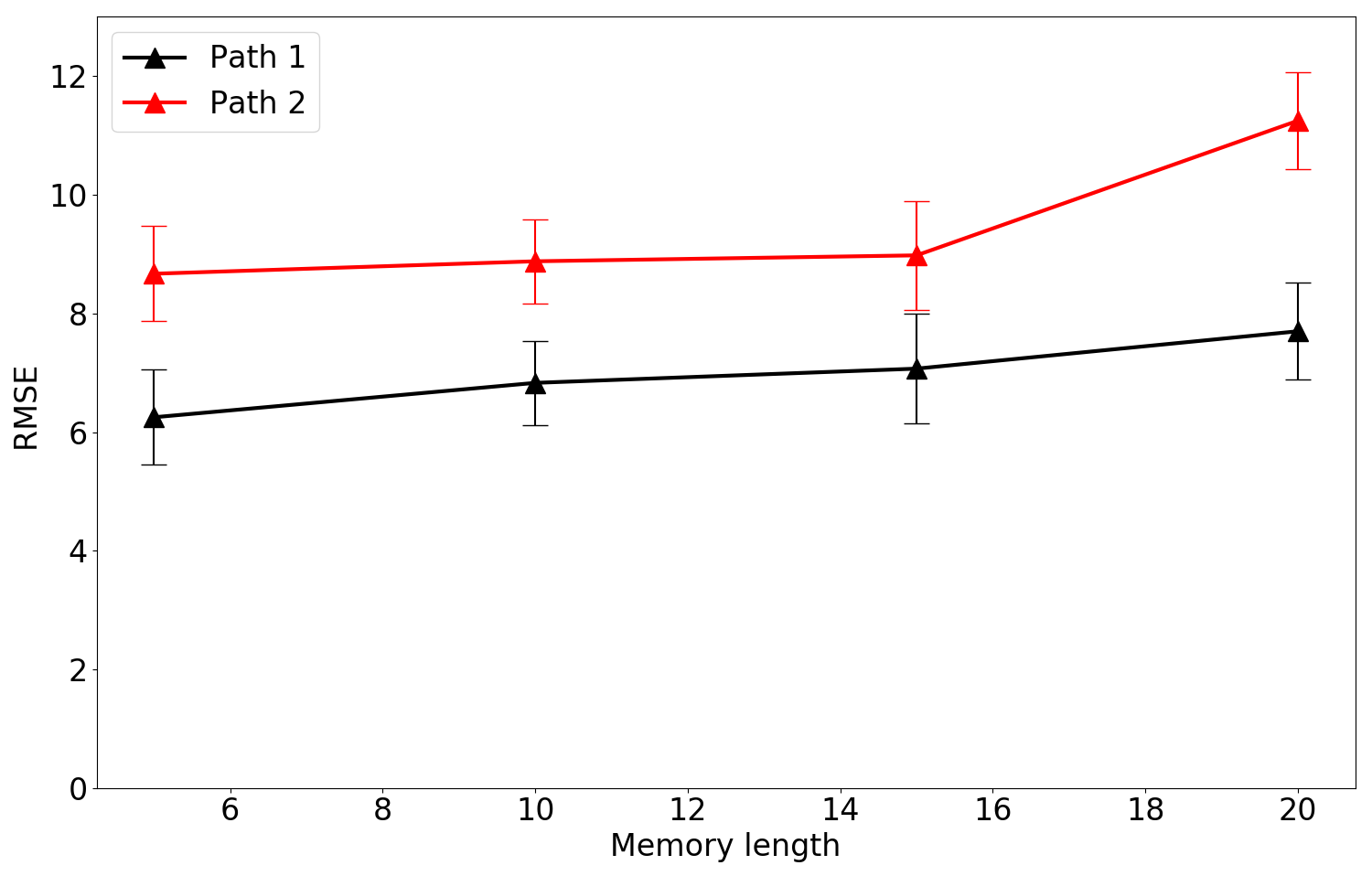}
\caption{Prediction results of different memory lengths in the RNN (the bars represent the standard deviations).}\label{Fig:memory_length}
\end{figure}

We also wan to identify the optimal hyperparameters for the CMDRNN model, i.e., the number of the mixture Gaussians for the MDN sub-model and the memory length for the RNN sub-model. From the results illustrated in Fig. \ref{Fig:mixture_number} and Fig. \ref{Fig:memory_length}, we can see the optimal mixture number is $30$ and the optimal memory length is $5$.   

Finally, we conduct several experiments to compare our model with other existing methods. The purposes of the experiments are indicated as follows: 

\begin{itemize}
    \item Comparing feature detectors: RNN, RNN+MDN, AE + RNN + MDN, CMDRNN;
    \item Comparing regressors: RNN, CNN+RNN and CMDRNN;
    \item Comparing RNN variants: CMDRNN, CMDLSTM and CMDGRU.
\end{itemize}

We use three baseline models, the k-NN, the decision tree and the random forests (which are not sequential models) ~\cite{rojo2019machine}. We run the algorithms multiple times with random initialisation. The results shown in Table~\ref{Tab:CMDRNN_result}, we can see that, compared to the baseline models, our proposed models outperform other approaches, and especially the CMDGRU model has the best performances. 

\begin{table*}
\centering
\caption{Root mean squared errors of the path prediction results (meter).}
\label{Tab:CMDRNN_result}
    \begin{subtable}{\textwidth}
    \small
    \centering
    \caption{Results of the baseline models.}
    \begin{tabular}{|c|c|c|c|}
    \hline
     Path & k-NN & DT & RF \\
    \hline
    Path 1 & $7.44 \pm 0.00$ & $8.78\pm 0.76$  & $7.25\pm 0.25$\\
    Path 2 & $8.02 \pm 0.00$ & $20.94\pm 1.52$ & $9.60\pm 0.75$\\
    \hline
    \end{tabular}
    \end{subtable}
    
    \vskip 0.3cm 
    
    \begin{subtable}{\textwidth}
    \small
    \centering
    \caption{Results of the sequential prediction models.}
    \begin{tabular}{|c|c|c|c|c|c|c|c|}
    \hline
     Path & RNN & CNN+RNN & RNN+MDN & AE+RNN+MDN & CMDRNN & CMDLSTM & CMDGRU\\
    \hline
    Path 1 & $29.36 \pm 1.61 $ & $34.26\pm 3.04$ & $23.86\pm 5.50$ & $11.24 \pm 0.86$ & $8.26\pm 1.31$ & $7.38\pm 0.89$ & $6.25\pm 0.80$ \\
    Path 2 & $31.61\pm 0.74 $ & $36.75 \pm 6.17$ & $23.58\pm 2.29$ & $12.01 \pm 1.68$ & $10.17\pm 0.72$ & $9.26\pm 0.31$ & $ 8.67\pm 0.23$ \\
    \hline
    \end{tabular}
    \end{subtable}
\end{table*}

\subsection{Location Recognition via Semi-supervised Learning}

\subsubsection{Dataset Description}

For the validation dataset, we use the UJIIndoorLoc dataset \cite{torres2014ujiindoorloc} which is similar to the Tampere dataset. The input dimension of the UJIIndoorLoc dataset is $520$. The possible RSSI values of the detected WAPs range from $-100 $ dB to $0$ dB and the RSSI values of undetected WAPs are set to $100$. The coordinates are in longitudes and latitudes. The total instances number for the experiments is about $20000$. 

The units of the target values are meters. For pre-processing the data, we set the undetected values the into $0$ dB for the purpose of computational convenience, which does not represent the actual signal strength. We also remove the duplicate instances. The original target data are longitudes and latitudes with very large values. They are scaled in the experiments, thus the predicting results in Table~\ref{Tab:VAE_result} do not have units. We run the algorithms multiple times with random initialisation.

\subsubsection{VAE-based Model Implementation Details}

The VAE-based model consists of three sub-networks, an encoder, a decoder and a predictor. The implementation details of the VAE-based semi-supervised learning model are demonstrated in Table~\ref{Tab:VAE_Imp}.

\begin{table*}
\centering
\caption{VAE-based model implementation details.}
\label{Tab:VAE_Imp}
\begin{tabular}{|c|c|c|c|}
\hline
Sub-network & Layer & Hyperparameter & Activation function\\
\hline
Encoder & hidden layer & neuron number: 512;  & ReLU\\
Encoder & hidden layer & neuron number: 512; latent dimension: 5 & ReLU\\
Decoder & hidden layer & neuron number: 512 & ReLU \\
Predictor & hidden layer & neuron number: 512; dropout rate: 0.3  & ReLU \\
Predictor & hidden layer & neuron number: 512; dropout rate: 0.3  & ReLU \\
Predictor & hidden layer & neuron number: 512; dropout rate: 0.3  & ReLU \\
\hline

\multicolumn{4}{|c|}{Optimizer: Adam; learning rate: $10^{-3}$} \\

\hline
\end{tabular}
\end{table*}

\subsubsection{Experimental Results}

For the experimental set up, we use different portions of labeled data ranging from $2\%$ to $80\%$. We use the k-NN, the DT and the RF as the baseline models and use the GP, the MDN with $2$ mixtures, noted as MDN(2), the MDN with $5$ mixtures, noted as MDN(5), as comparisons. 

\begin{figure}
\centering
    \begin{subfigure}[t]{0.4\textwidth}
    \includegraphics[width= \linewidth]{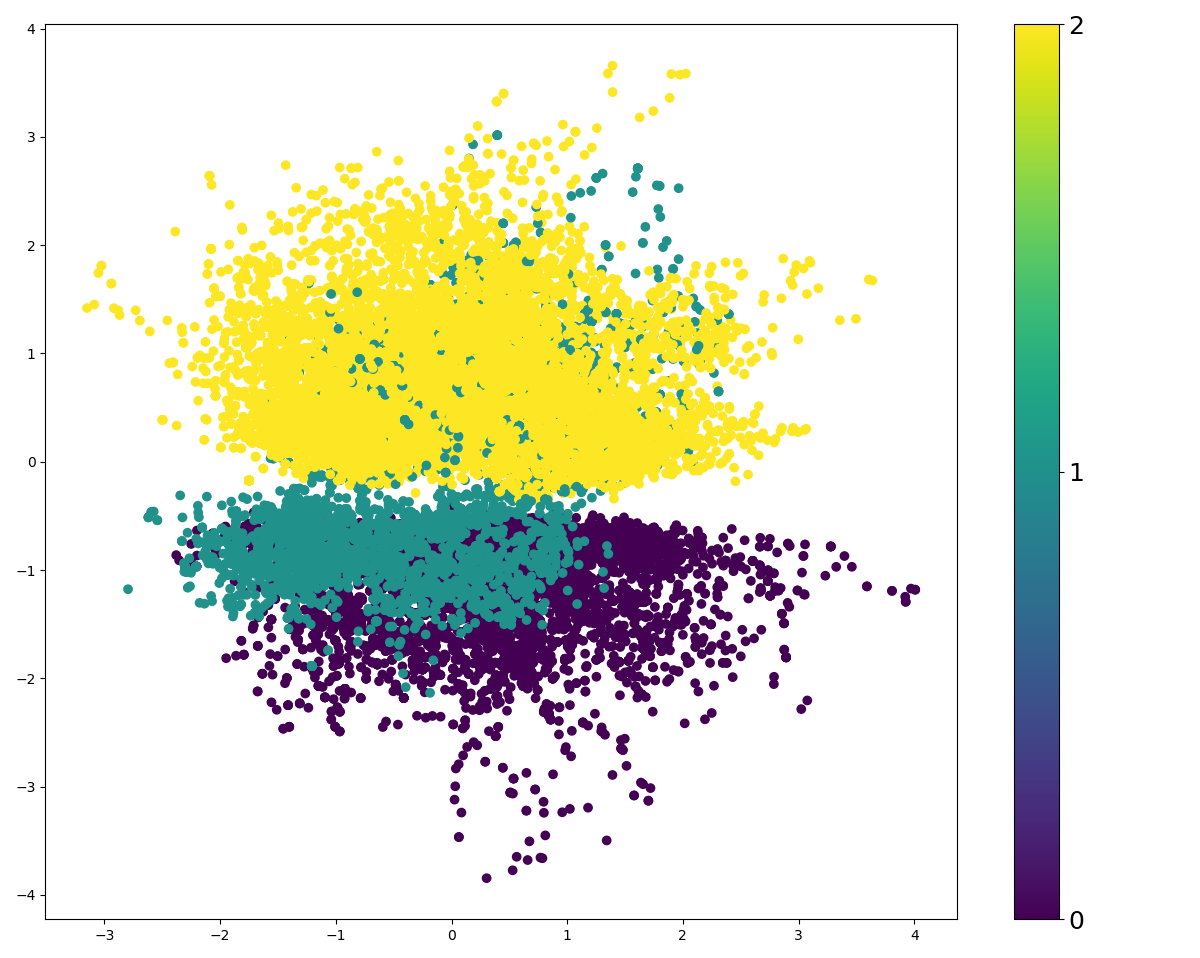}
    \caption{Latent variables labeled with the building IDs.}\label{Fig:Z_B}
    \end{subfigure}

    \begin{subfigure}[!t]{0.4\textwidth}
    \centering
    \includegraphics[width= \linewidth]{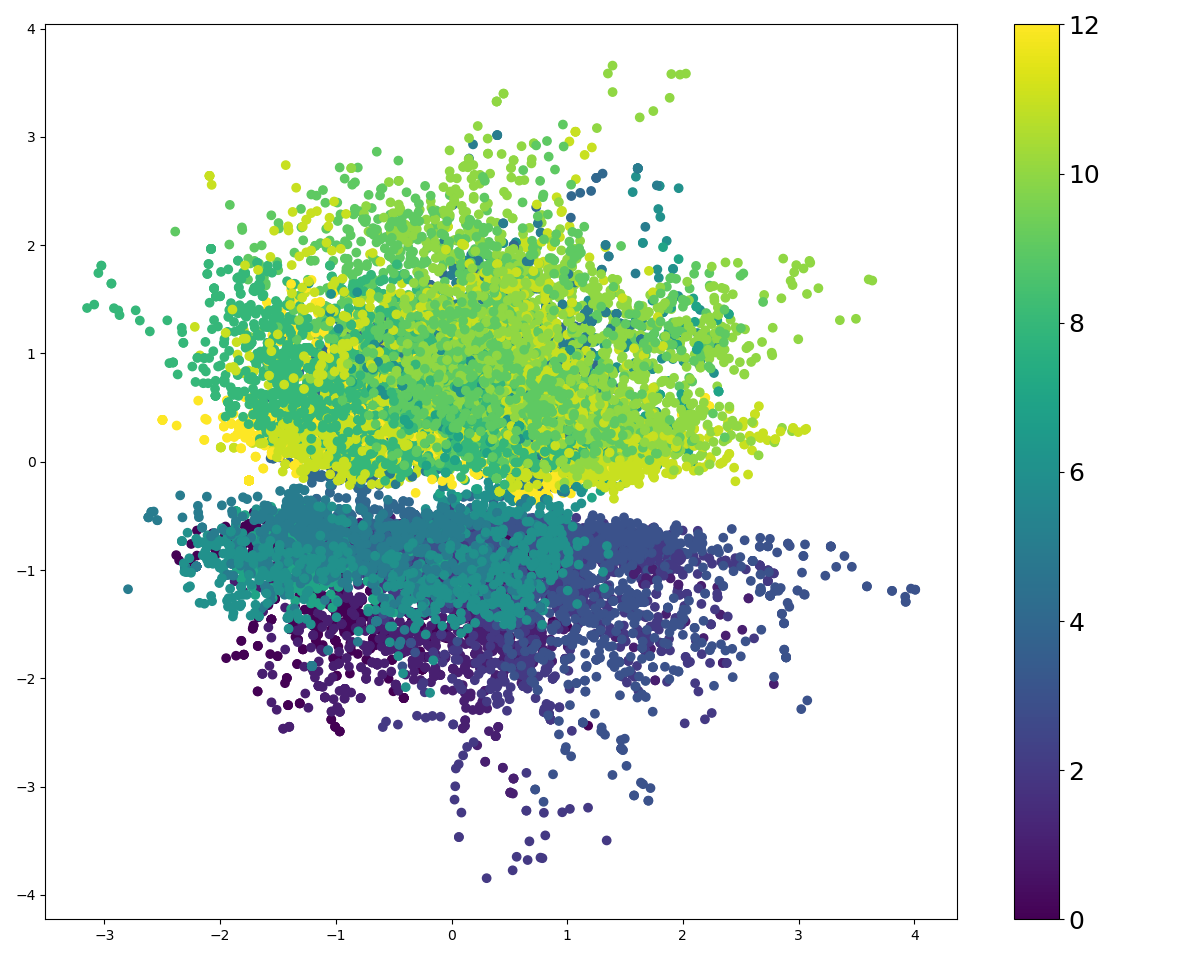}
    \caption{Latent variables labeled with the floor IDs.}\label{Fig:Z_F}
    \end{subfigure}
    
\caption{Latent variables with dimension of $5$, here shows the 2D projection.}\label{Fig:Latent}    
\end{figure}

Fig.~\ref{Fig:Latent} demonstrates the distribution of the latent variable $z$. The results verify our assumption that  the latent variable $z$ are related to both the input $x$ and output $y$.

\begin{figure}
    \centering
    \begin{subfigure}{0.2\textwidth}
        \centering
        \includegraphics[height=1.1in]{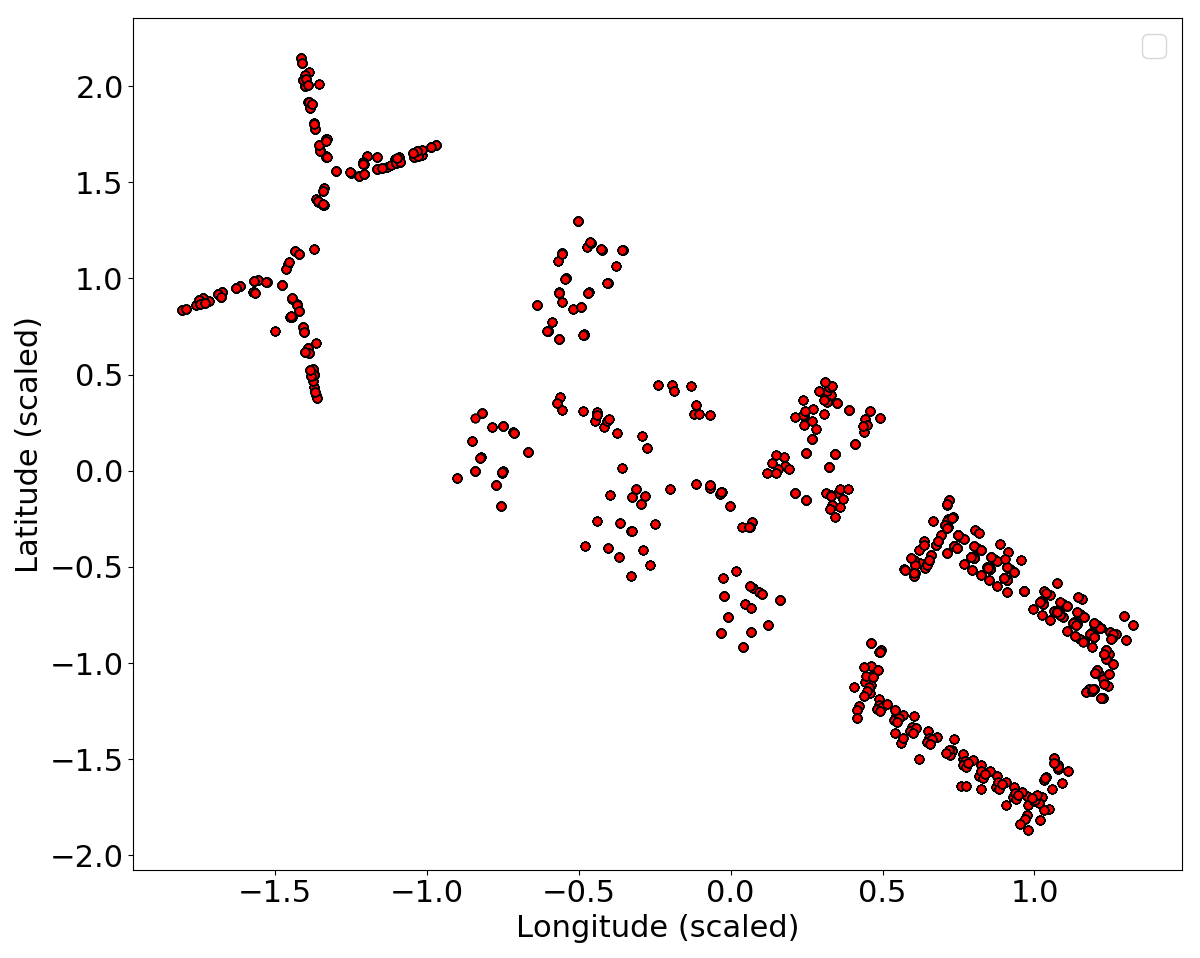}
        \caption{Ground truth.}
    \end{subfigure}%
    \!
    \begin{subfigure}{0.2\textwidth}
        \centering
        \includegraphics[height=1.1in]{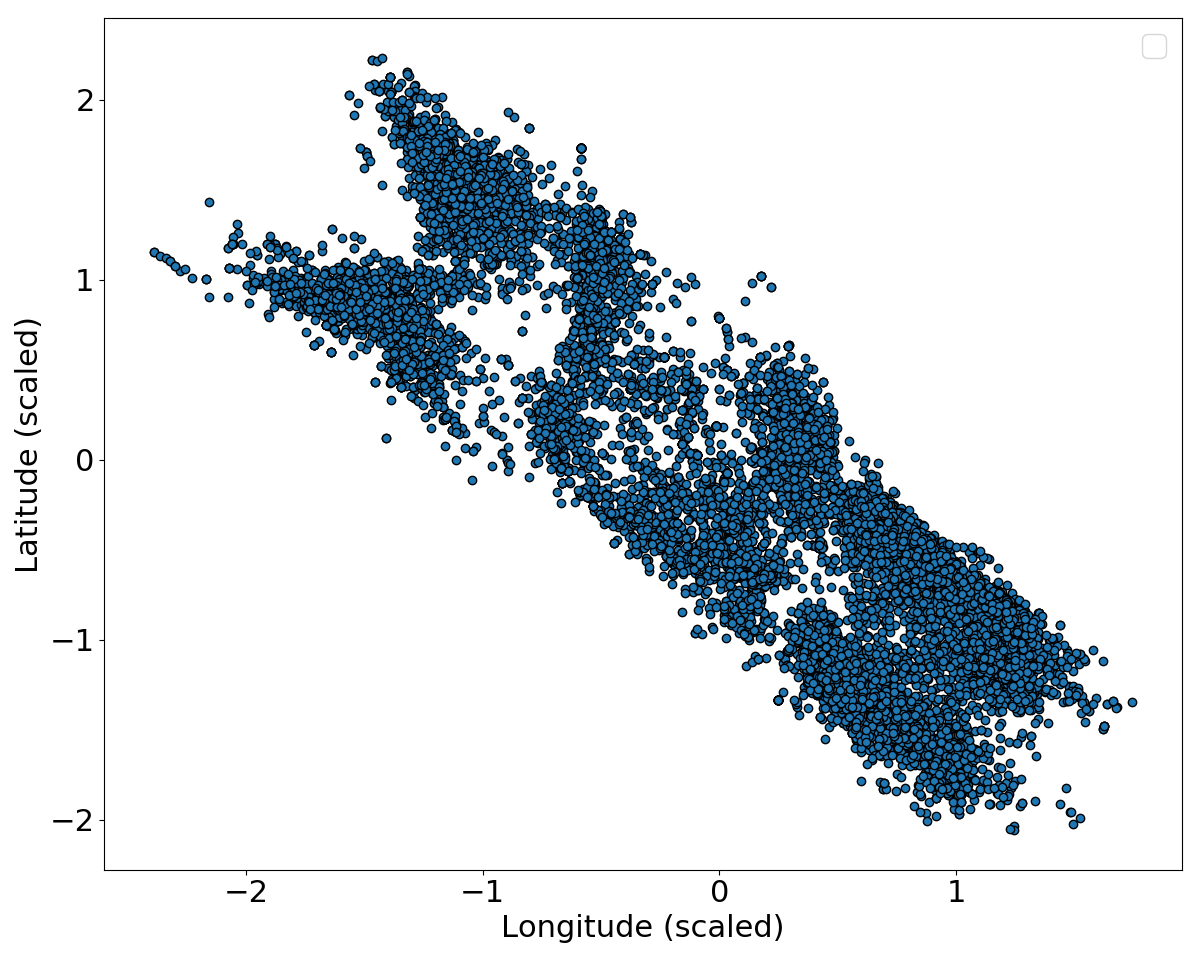}
        \subcaption{Labeled data: 2\%.}
    \end{subfigure}

    \begin{subfigure}{0.2\textwidth}
        \centering
        \includegraphics[height=1.1in]{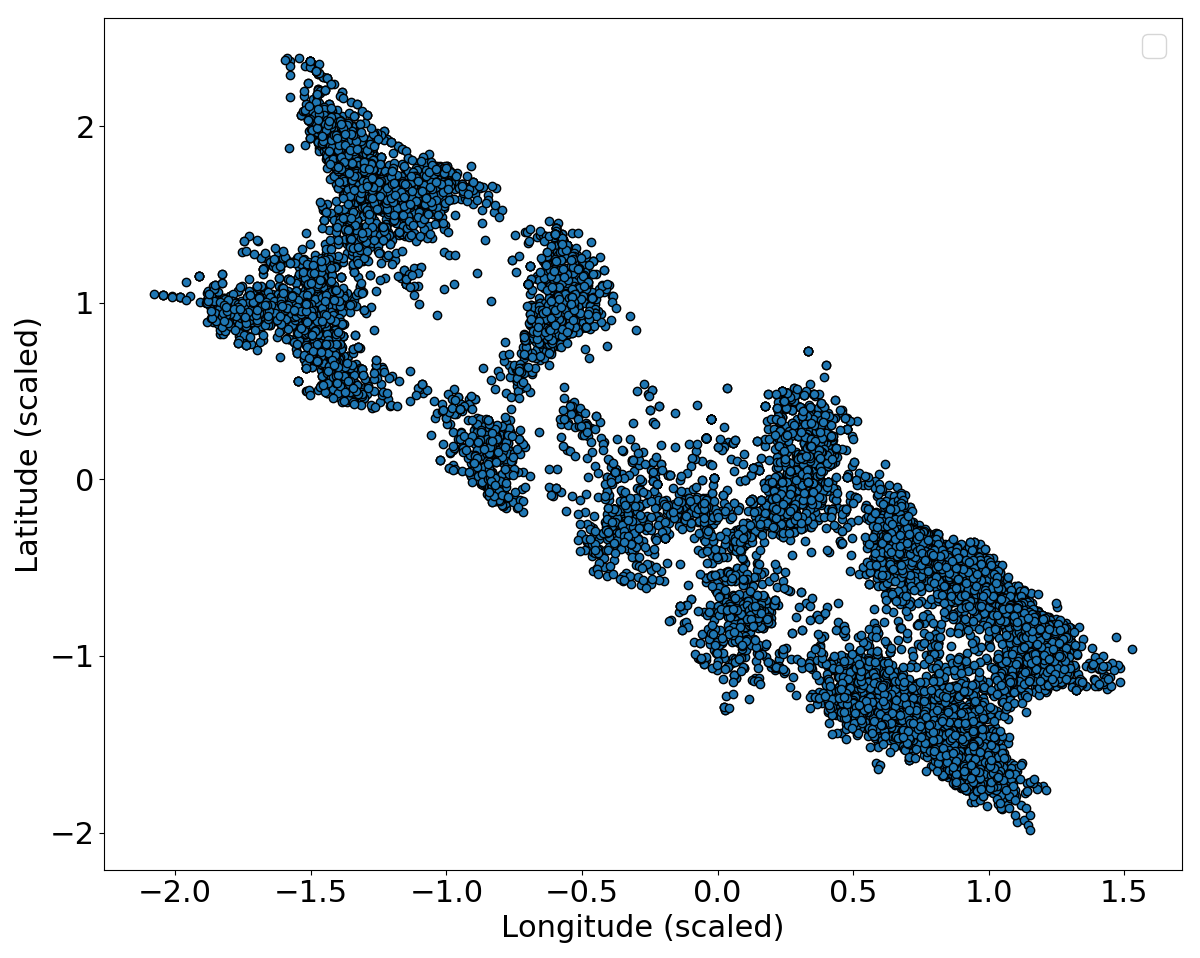}
        \subcaption{Labeled data: 5\%.}
    \end{subfigure}
     \! 
    \begin{subfigure}{0.2\textwidth}
        \centering
        \includegraphics[height=1.1in]{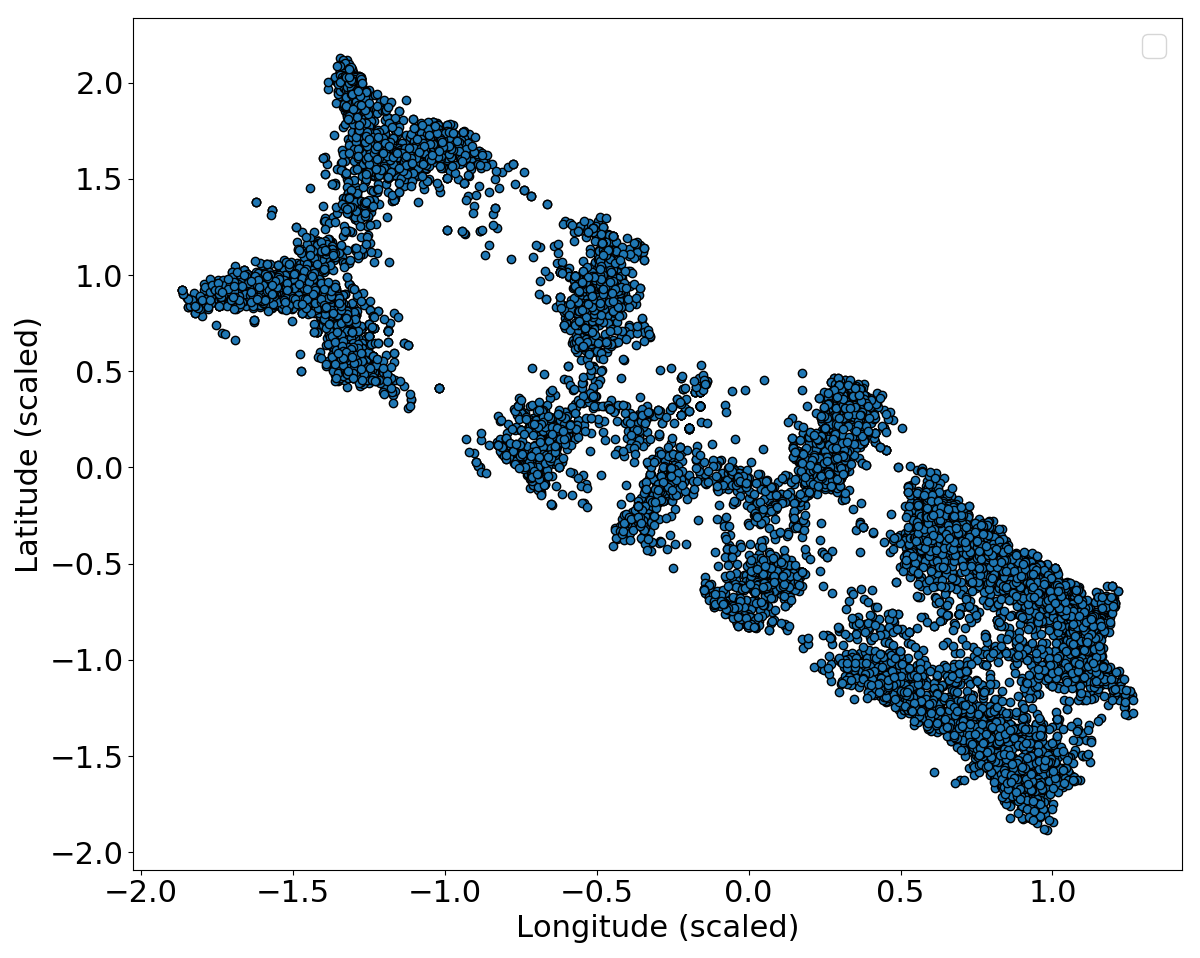}
        \subcaption{Labeled data: 10\%.}
    \end{subfigure}

    \begin{subfigure}{0.2\textwidth}
        \centering
        \includegraphics[height=1.1in]{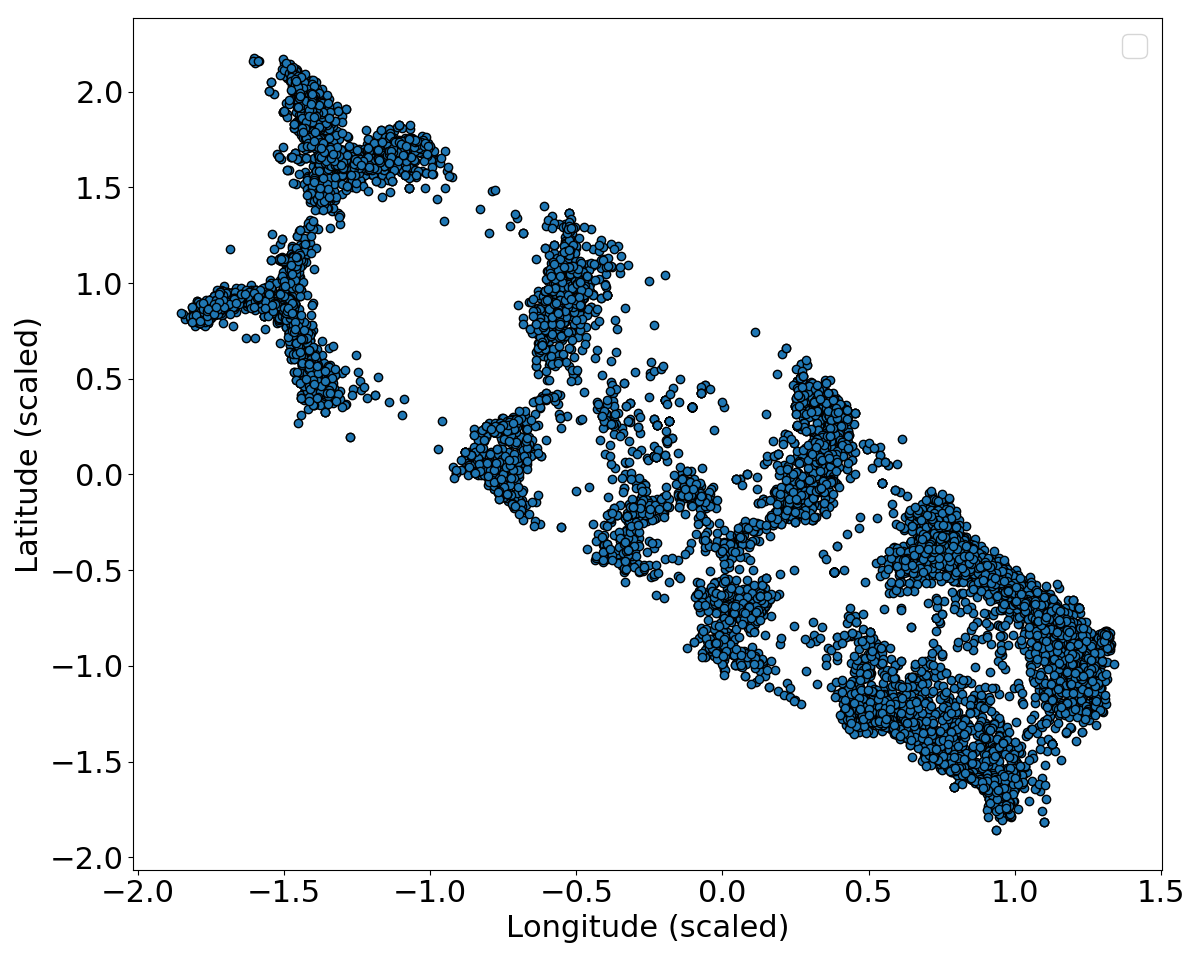}
        \subcaption{Labeled data: 20\%.}
    \end{subfigure}
     \! 
    \begin{subfigure}{0.2\textwidth}
        \centering
        \includegraphics[height=1.1in]{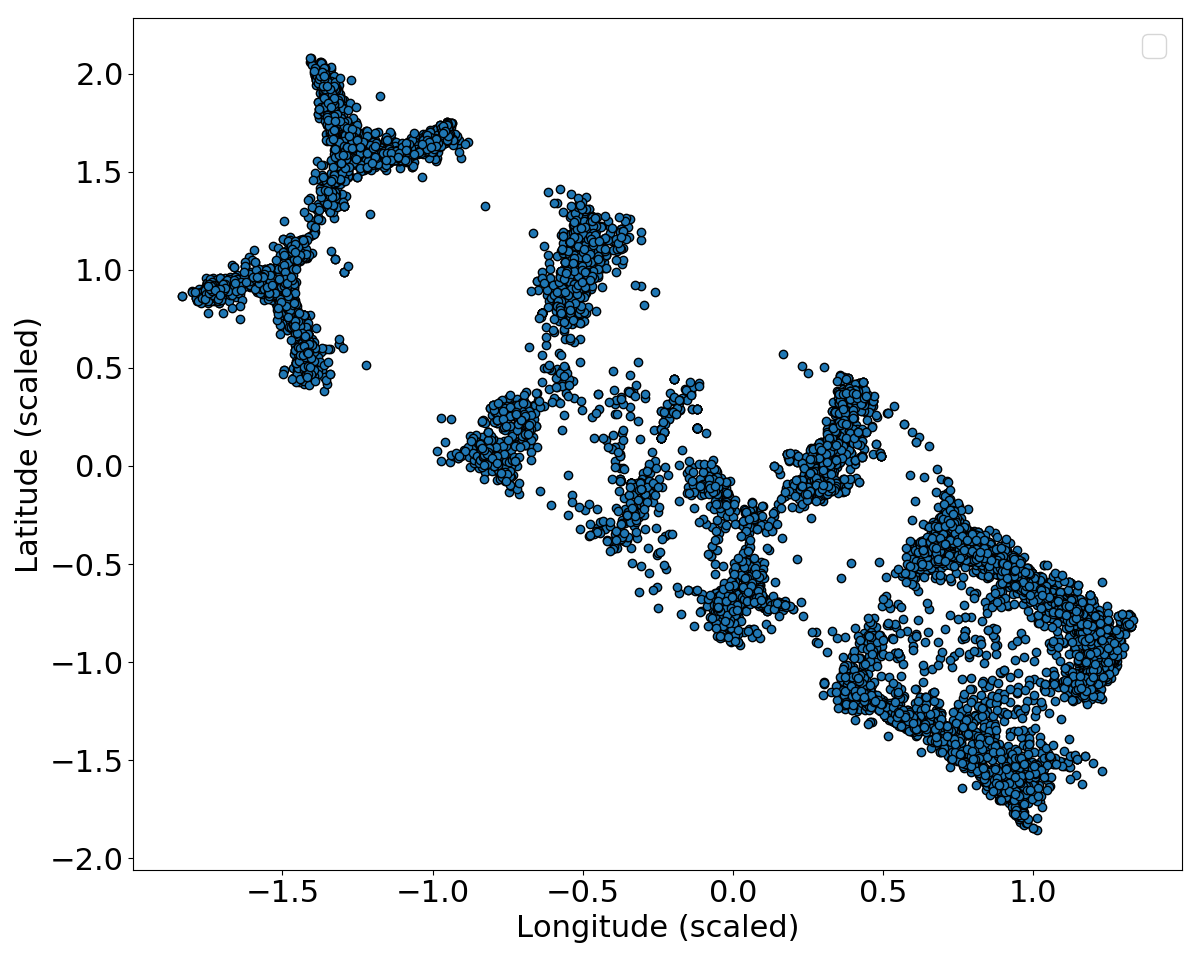}
        \subcaption{Labeled data: 30\%.}
    \end{subfigure}  
    
     \begin{subfigure}{0.2\textwidth}
        \centering
        \includegraphics[height=1.1in]{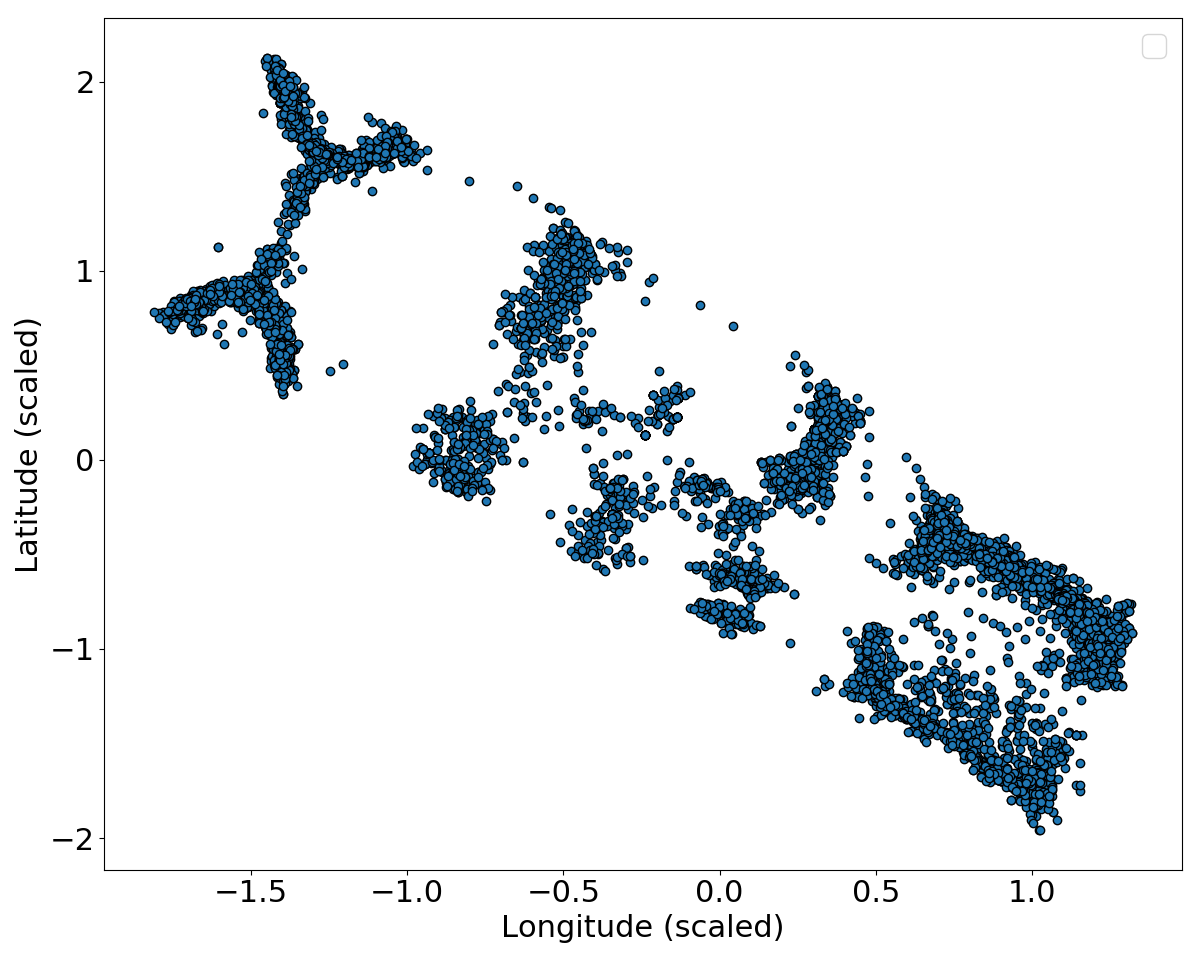}
        \subcaption{Labeled data: 50\%.}
    \end{subfigure}
     \! 
    \begin{subfigure}{0.2\textwidth}
        \centering
        \includegraphics[height=1.1in]{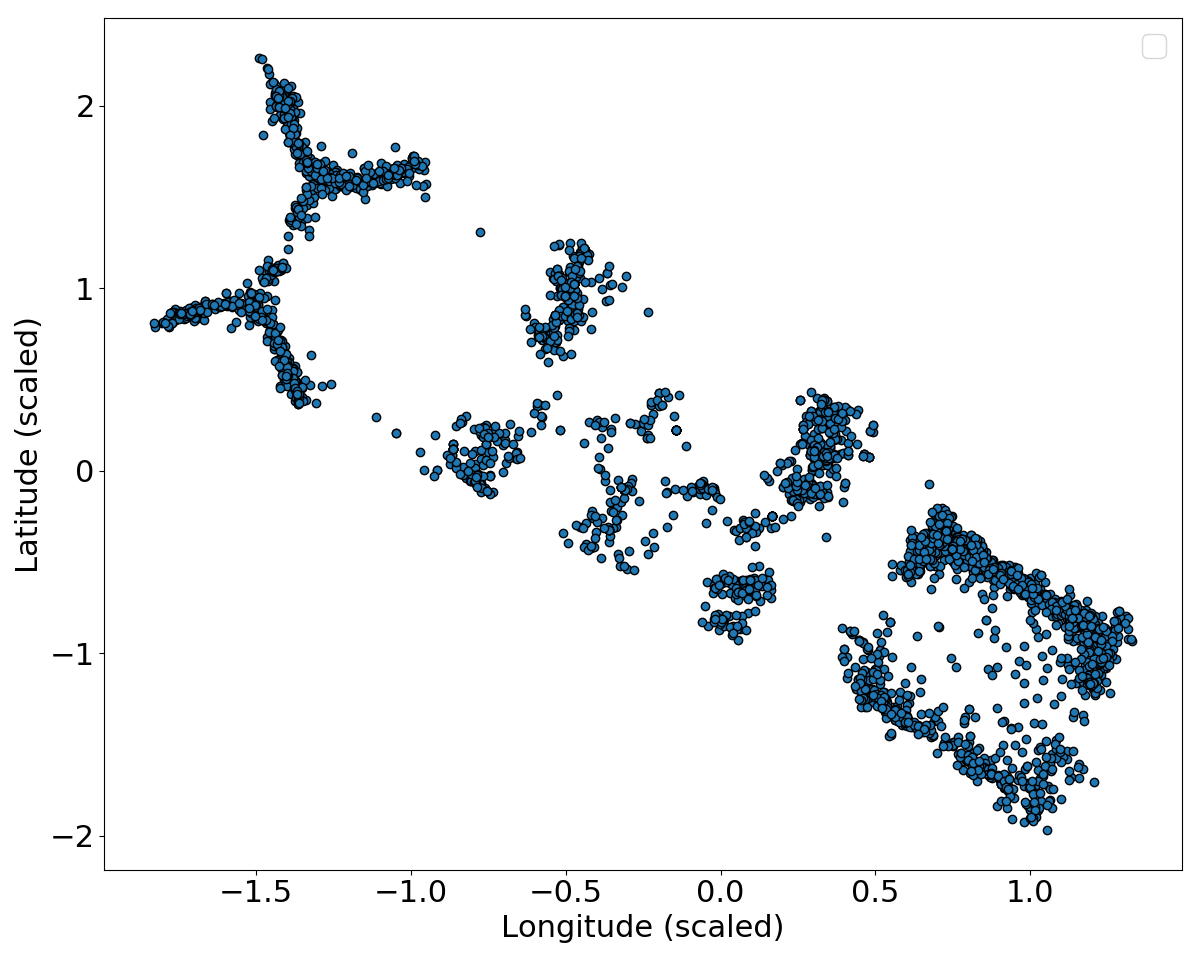}
        \subcaption{Labeled data: 80\%.}
    \end{subfigure} 

    \caption{Test results of the M2 model.}
    \label{Fig:VAE_result}
\end{figure}

Fig.~\ref{Fig:VAE_result} shows the results obtained by the M2 model with respect to different amounts of the labeled data. We can see that the accuracy of the testing results increases as the amount of the labeled data increases in the supervised learning procedure.     

\begin{table*}
\centering
\caption{Root mean squared errors of testing results with respect to different portions of labeled data.}
\label{Tab:VAE_result}
\begin{subtable}{\textwidth}
\small
\centering
\caption{Results of the baseline models.}
\begin{tabular}{|c|c|c|c|}
\hline
 Labeled data & k-NN & DT & RF\\
\hline
2\%  & $0.202\pm3\text{e-}2$ & $0.349\pm5\text{e-}2$ & $0.234\pm3\text{e-}2$  \\

5\%  & $0.177\pm1\text{e-}2$ & $0.254\pm2\text{e-}2$ & $0.184\pm2\text{e-}2$  \\

10\% & $0.156\pm6\text{e-}3$ & $0.201\pm2\text{e-}2$ & $0.138\pm6\text{e-}3$  \\

20\% & $0.120\pm4\text{e-}3$ & $0.161\pm9\text{e-}3$ & $0.112\pm5\text{e-}4$  \\

30\% & $0.104\pm2\text{e-}3$ & $0.140\pm4\text{e-}2$ & $0.100\pm2\text{e-}3$  \\

50\% & $0.100\pm1\text{e-}2$ & $0.126\pm2\text{e-}3$ & $0.091\pm2\text{e-}3$  \\

80\% & $0.092\pm7\text{e-}3$ & $0.112\pm3\text{e-}3$ & $0.087\pm3\text{e-}3$  \\
\hline
\end{tabular}
\end{subtable}

    \vskip 0.3cm 
    
\begin{subtable}{\textwidth}
\small
\centering
\caption{Results of the comparison models.}
\begin{tabular}{|c|c|c|c|c|c|}
\hline
 Labeled data & GP & MDN(2) & MDN(5) & M1 & M2\\
\hline
2\%  & $0.565\pm 3\text{e-}2$ & $0.161\pm5\text{e-}3$ & $0.159\pm4\text{e-}3$ & $0.175\pm6\text{e-}3$ & $0.166\pm8\text{e-}3$\\

5\%  & $0.313\pm 2\text{e-}2$ & $0.139\pm4\text{e-}3$ & $0.143\pm3\text{e-}3$ & $0.133\pm3\text{e-}3$ & $0.123\pm4\text{e-}3$\\

10\% & $0.275\pm 3\text{e-}3$ & $0.120\pm6\text{e-}3$ & $0.129\pm1\text{e-}3$ & $0.105\pm2\text{e-}4$ & $0.106\pm5\text{e-}3$\\

20\% & $0.262\pm 8\text{e-}4$ & $0.107\pm6\text{e-}3$ & $0.105\pm4\text{e-}3$ & $0.093\pm2\text{e-}3$ & $0.093\pm2\text{e-}3$\\

30\% & $0.258\pm 2\text{e-}3$ & $0.102\pm5\text{e-}3$ & $0.105\pm4\text{e-}3$ & $0.086\pm3\text{e-}3$ & $0.087\pm2\text{e-}3$\\

50\% & $0.253\pm 2\text{e-}3$ & $0.101\pm7\text{e-}3$ & $0.097\pm4\text{e-}3$ & $0.080\pm9\text{e-}4$ & $0.083\pm4\text{e-}3$\\

80\% & $0.253\pm 2\text{e-}3$ & $0.098\pm7\text{e-}3$ & $0.103\pm3\text{e-}3$ & $0.077\pm4\text{e-}3$ & $0.079\pm3\text{e-}3$\\
\hline
\end{tabular}
\end{subtable}
\end{table*}

Table~\ref{Tab:VAE_result} shows the results obtained by different methods with respect to different amounts of the labeled data. From the results, we can see that the proposed models outperform the baseline model proposed in~\cite{rojo2019machine}. Moreover, M1 and M2 can provide satisfying results even when the labeled data are scarce. The predicting accuracy is improved when the labeled data increases. In contrast with other existing machine learning and deep learning methods, the proposed models have better performance. In practice, we also find that the proposed models, compared to other methods, have such following advantages:  

\begin{itemize}
    \item Compared to the GP, the VAE-based models are less computationally expensive when the training dataset is large;
    \item Compared to the MDNs, the VAE-based models are more computationally stable during the learning process.
\end{itemize}

\section{Conclusions and Perspectives}

\label{sec:Conclusions}

In this paper, we aim to tackle the WiFi fingerprint-based indoor positioning problem. The first task in our work is predicting next user location with WiFi fingerprints. By contrast with existing approaches, our solution is a hybrid deep learning model. The proposed model is composed of three deep neural networks, a CNN, a RNN and a MDN. This unique deep architecture combines all the potentials of three deep learning models, which enables us to predict user location with relatively high accuracy. 

The second task in our work is recognizing accurate user location via semi-supervised learning. To solve this problem, we propose a VAE-based semi-supervised learning model. Our model employs a VAE model to carry out the unsupervised learning procedure so as to learn a latent distribution of the original input. Afterwards, in order to use the learnt distribution to compute the user coordinates, we devise two different predictors, one is deterministic, the other is stochastic.  

Finally, we test our models on two real-world datasets. From the results, we can see that the CMDRNN model has better performance than conventional machine learning methods and conventional RNN-based models for indoor next location prediction. The VAE-based model outperform other conventional machine learning and deep learning methods for accurate location recognition. For the future work, we plan to explore other advanced deep learning methods to develop new applications.

\bibliographystyle{unsrt}

\bibliography{cas-refs}

\begin{thebibliography}{10}

\bibitem{bahl2000radar}
Paramvir Bahl and Venkata~N Padmanabhan.
\newblock Radar: An in-building rf-based user location and tracking system.
\newblock In {\em Proceedings IEEE INFOCOM 2000. Conference on Computer
  Communications. Nineteenth Annual Joint Conference of the IEEE Computer and
  Communications Societies (Cat. No. 00CH37064)}, volume~2, pages 775--784.
  Ieee, 2000.

\bibitem{youssef2008horus}
Moustafa Youssef and Ashok Agrawala.
\newblock The horus location determination system.
\newblock {\em Wireless Networks}, 14(3):357--374, 2008.

\bibitem{cho2016exploiting}
Sung-Bae Cho.
\newblock Exploiting machine learning techniques for location recognition and
  prediction with smartphone logs.
\newblock {\em Neurocomputing}, 176:98--106, 2016.

\bibitem{yu2017modeling}
Chen Yu, Yang Liu, Dezhong Yao, Laurence~T Yang, Hai Jin, Hanhua Chen, and
  Qiang Ding.
\newblock Modeling user activity patterns for next-place prediction.
\newblock {\em IEEE Systems Journal}, 11(2):1060--1071, 2017.

\bibitem{hoang2019recurrent}
Minh~Tu Hoang, Brosnan Yuen, Xiaodai Dong, Tao Lu, Robert Westendorp, and
  Kishore Reddy.
\newblock Recurrent neural networks for accurate rssi indoor localization.
\newblock {\em IEEE Internet of Things Journal}, 6(6):10639--10651, 2019.

\bibitem{bozkurt2015comparative}
Sinem Bozkurt, Gulin Elibol, Serkan Gunal, and Ugur Yayan.
\newblock A comparative study on machine learning algorithms for indoor
  positioning.
\newblock In {\em 2015 International Symposium on Innovations in Intelligent
  SysTems and Applications (INISTA)}, pages 1--8. IEEE, 2015.

\bibitem{cramariuc2016clustering}
Andrei Cramariuc, Heikki Huttunen, and Elena~Simona Lohan.
\newblock Clustering benefits in mobile-centric wifi positioning in multi-floor
  buildings.
\newblock In {\em 2016 International Conference on Localization and GNSS
  (ICL-GNSS)}, pages 1--6. IEEE, 2016.

\bibitem{torres2015comprehensive}
Joaqu{\'\i}n Torres-Sospedra, Ra{\'u}l Montoliu, Sergio Trilles, {\'O}scar
  Belmonte, and Joaqu{\'\i}n Huerta.
\newblock Comprehensive analysis of distance and similarity measures for wi-fi
  fingerprinting indoor positioning systems.
\newblock {\em Expert Systems with Applications}, 42(23):9263--9278, 2015.

\bibitem{rasmussen2003gaussian}
Carl~Edward Rasmussen.
\newblock Gaussian processes in machine learning.
\newblock In {\em Summer School on Machine Learning}, pages 63--71. Springer,
  2003.

\bibitem{ferris2007wifi}
Brian Ferris, Dieter Fox, and Neil~D Lawrence.
\newblock Wifi-slam using gaussian process latent variable models.
\newblock In {\em IJCAI}, volume~7, pages 2480--2485, 2007.

\bibitem{hahnel2006gaussian}
Brian Ferris~Dirk H{\"a}hnel and Dieter Fox.
\newblock Gaussian processes for signal strength-based location estimation.
\newblock In {\em Proceeding of robotics: science and systems}, 2006.

\bibitem{yiu2015gaussian}
Simon Yiu and Kai Yang.
\newblock Gaussian process assisted fingerprinting localization.
\newblock {\em IEEE Internet of Things Journal}, 3(5):683--690, 2015.

\bibitem{yang2019online}
Tao Yang, Cindy Cappelle, Yassine Ruichek, and Mohammed El~Bagdouri.
\newblock Online multi-object tracking combining optical flow and compressive
  tracking in markov decision process.
\newblock {\em Journal of Visual Communication and Image Representation},
  58:178--186, 2019.

\bibitem{krogh2001predicting}
Anders Krogh, Bjo{\`E}rn Larsson, Gunnar Von~Heijne, and Erik~LL Sonnhammer.
\newblock Predicting transmembrane protein topology with a hidden markov model:
  application to complete genomes.
\newblock {\em Journal of molecular biology}, 305(3):567--580, 2001.

\bibitem{lecun1998gradient}
Yann LeCun, L{\'e}on Bottou, Yoshua Bengio, Patrick Haffner, et~al.
\newblock Gradient-based learning applied to document recognition.
\newblock {\em Proceedings of the IEEE}, 86(11):2278--2324, 1998.

\bibitem{hinton2006reducing}
Geoffrey~E Hinton and Ruslan~R Salakhutdinov.
\newblock Reducing the dimensionality of data with neural networks.
\newblock {\em science}, 313(5786):504--507, 2006.

\bibitem{elman1990finding}
Jeffrey~L Elman.
\newblock Finding structure in time.
\newblock {\em Cognitive science}, 14(2):179--211, 1990.

\bibitem{ibrahim2018cnn}
Mai Ibrahim, Marwan Torki, and Mustafa ElNainay.
\newblock Cnn based indoor localization using rss time-series.
\newblock In {\em 2018 IEEE Symposium on Computers and Communications (ISCC)},
  pages 01044--01049. IEEE, 2018.

\bibitem{ayyalasomayajula2020deep}
Roshan Ayyalasomayajula, Aditya Arun, Chenfeng Wu, Sanatan Sharma,
  Abhishek~Rajkumar Sethi, Deepak Vasisht, and Dinesh Bharadia.
\newblock Deep learning based wireless localization for indoor navigation.
\newblock In {\em Proceedings of the 26th Annual International Conference on
  Mobile Computing and Networking}, pages 1--14, 2020.

\bibitem{nowicki2017low}
Micha{\l} Nowicki and Jan Wietrzykowski.
\newblock Low-effort place recognition with wifi fingerprints using deep
  learning.
\newblock In {\em International Conference Automation}, pages 575--584.
  Springer, 2017.

\bibitem{song2019novel}
Xudong Song, Xiaochen Fan, Chaocan Xiang, Qianwen Ye, Leyu Liu, Zumin Wang,
  Xiangjian He, Ning Yang, and Gengfa Fang.
\newblock A novel convolutional neural network based indoor localization
  framework with wifi fingerprinting.
\newblock {\em IEEE Access}, 7:110698--110709, 2019.

\bibitem{kim2018scalable}
Kyeong~Soo Kim, Sanghyuk Lee, and Kaizhu Huang.
\newblock A scalable deep neural network architecture for multi-building and
  multi-floor indoor localization based on wi-fi fingerprinting.
\newblock {\em Big Data Analytics}, 3(1):4, 2018.

\bibitem{bishop2006pattern}
Christopher~M Bishop.
\newblock {\em Pattern recognition and machine learning}.
\newblock Springer Science+ Business Media, 2006.

\bibitem{bishop1994mixture}
{Bishop, Christopher M}.
\newblock Mixture density networks.
\newblock 1994.

\bibitem{hernandez2015probabilistic}
Jos{\'e}~Miguel Hern{\'a}ndez-Lobato and Ryan Adams.
\newblock Probabilistic backpropagation for scalable learning of bayesian
  neural networks.
\newblock In {\em International Conference on Machine Learning}, pages
  1861--1869, 2015.

\bibitem{kingma2014auto}
Diederik~P. Kingma and Max Welling.
\newblock Auto-encoding variational bayes.
\newblock In Yoshua Bengio and Yann LeCun, editors, {\em Proceedings of the 3rd
  International Conference on Learning Representations (ICLR)}, 2014.

\bibitem{hochreiter1997long}
Sepp Hochreiter and J{\"u}rgen Schmidhuber.
\newblock Long short-term memory.
\newblock {\em Neural computation}, 9(8):1735--1780, 1997.

\bibitem{chung2014empirical}
Junyoung Chung, Caglar Gulcehre, Kyunghyun Cho, and Yoshua Bengio.
\newblock Empirical evaluation of gated recurrent neural networks on sequence
  modeling.
\newblock In {\em NIPS 2014 Workshop on Deep Learning, December 2014}, 2014.

\bibitem{lohan2017wi}
Elena~Simona Lohan, Joaqu{\'\i}n Torres-Sospedra, Helena Lepp{\"a}koski,
  Philipp Richter, Zhe Peng, and Joaqu{\'\i}n Huerta.
\newblock Wi-fi crowdsourced fingerprinting dataset for indoor positioning.
\newblock {\em Data}, 2(4):32, 2017.

\bibitem{tieleman2012lecture}
Tijmen Tieleman and Geoffrey Hinton.
\newblock Lecture 6.5-rmsprop: Divide the gradient by a running average of its
  recent magnitude.
\newblock {\em COURSERA: Neural networks for machine learning}, 4(2):26--31,
  2012.

\bibitem{martin2017wasserstein}
SC~Martin~Arjovsky and Leon Bottou.
\newblock Wasserstein generative adversarial networks.
\newblock In {\em Proceedings of the 34 th International Conference on Machine
  Learning, Sydney, Australia}, 2017.

\bibitem{kingma2015adam}
Diederik~P. Kingma and Jimmy Ba.
\newblock Adam: A method for stochastic optimization.
\newblock In Yoshua Bengio and Yann LeCun, editors, {\em 3rd International
  Conference on Learning Representations (ICLR)}, 2015.

\bibitem{rojo2019machine}
Jordi Rojo, Germ{\'a}n~Mart{\'\i}n Mendoza-Silva, Gabriel~Ristow Cidral, Jorma
  Laiapea, Gerardo Parrello, Arnau Sim{\'o}, Laura Stupin, Deniz Minican,
  Mar{\'\i}a Farr{\'e}s, Carmen Corval{\'a}n, et~al.
\newblock Machine learning applied to wi-fi fingerprinting: The experiences of
  the ubiqum challenge.
\newblock In {\em 2019 International Conference on Indoor Positioning and
  Indoor Navigation (IPIN)}, pages 1--8. IEEE, 2019.

\bibitem{torres2014ujiindoorloc}
Joaqu{\'\i}n Torres-Sospedra, Ra{\'u}l Montoliu, Adolfo Mart{\'\i}nez-Us{\'o},
  Joan~P Avariento, Tom{\'a}s~J Arnau, Mauri Benedito-Bordonau, and
  Joaqu{\'\i}n Huerta.
\newblock Ujiindoorloc: A new multi-building and multi-floor database for wlan
  fingerprint-based indoor localization problems.
\newblock In {\em 2014 international conference on indoor positioning and
  indoor navigation (IPIN)}, pages 261--270. IEEE, 2014.

\end{thebibliography}

\end{document}